\documentclass[journal,comsoc]{IEEEtran} 
\usepackage{amsmath}
\usepackage{newtxmath} 
\usepackage{color} 
\usepackage{url}
\usepackage{graphicx}
\usepackage{bm}
\usepackage{bbm, dsfont}
\usepackage{xcolor, soul}
\usepackage{varwidth}
\usepackage{cite}
\usepackage{float, booktabs}
\usepackage{subfigure}
\usepackage{multirow}
\usepackage{flushend}

\makeatletter

\ifCLASSINFOpdf
\else
\fi
\makeatother

\begin{document}

\title{A Multi-task Neural Approach for Emotion Attribution, Classification and Summarization} \author{Guoyun~Tu$^\dagger$,
Yanwei~Fu$^{\dagger}$, Boyang~Li,  Jiarui~Gao,  Yu-Gang~Jiang and Xiangyang~Xue
\IEEEcompsocitemizethanks{ \IEEEcompsocthanksitem Guoyun~Tu, Yanwei~Fu,
Jiarui~Gao, Yu-Gang~Jiang and Xiangyang~Xue are with Fudan University, Shanghai,
China. $\dagger$ indicates equal contribution.
\IEEEcompsocthanksitem Boyang Li is with the Big Data Lab of Baidu Research, Sunnyvale, California, US 94089. Email: boyangli@baidu.com.
\IEEEcompsocthanksitem Yanwei Fu  is with the School of  Data Science, Fudan University, Shanghai, China. Email:yanweifu@fudan.edu.cn.
\IEEEcompsocthanksitem  Y.-G. Jiang (corresponding author) is with School of Computer Science, Fudan University, and Jilian Technology Group (Video++), Shanghai, China, 200082. Email: ygj@fudan.edu.cn.
 } \thanks{} 
} 
\maketitle

\begin{abstract} Emotional content is a crucial ingredient in user-generated videos. However, the sparsity of emotional expressions in the videos poses an obstacle to visual emotion analysis. In this paper, we propose a new neural approach, Bi-stream Emotion Attribution-Classification Network (BEAC-Net), to solve three related emotion analysis tasks: emotion recognition, emotion attribution, and emotion-oriented summarization, in a single integrated framework. BEAC-Net has two major constituents, an attribution network and a classification network. The attribution network extracts the main emotional segment that classification should focus on in order to mitigate the sparsity issue. The classification network utilizes both the extracted segment and the original video in a bi-stream architecture. We contribute a new dataset for the emotion attribution task with human-annotated ground-truth labels for emotion segments. Experiments on two video datasets demonstrate superior performance of the proposed framework and the complementary nature of the dual classification streams. \end{abstract}

\section{Introduction}

The explosive growth of user-generated video has created great demand for computational understanding of visual data and attracted significant research attention in the multimedia community. Research from the past few decades shows the cognitive processes responsible for emotion appraisals and coping play important roles in human cognition \cite{Damasio1994,Clore2009}. It follows that computational understanding of emotional content in video will help predict how human audience will interact with video content and help answer questions like:
\begin{itemize}
\item Will a video recently posted on social media go viral in the next few hours \cite{Guadagno2013}?
\item  Will a commercial break disrupt the emotion of the video and ruin the viewing experience \cite{CAVVA}?
\end{itemize}  As an illustrative example, a food commercial probably should not accompany a video that elicits the emotion of disgust. Proper commercial placement would benefit from the precise location of emotions, in addition to identifying the overall emotion. 

Significant successes have been achieved on the problem of video understanding, such as the recognition of activities \cite{Ikizler2012,Xu2017} and participants \cite{Somandepalli2017}.
Nevertheless, computational recognition and structured understanding of
video emotion remains largely an open problem.
In this paper, we focus on the emotion perceived by the audience. 
The emotion may be expressed by facial expressions, event sequences (e.g., a wedding ceremony), nonverbal language, or even just abstract shapes and colors. This differs from work that focus on one particular channel such as the human face\cite{Joho2009,Zhao2011,Zhen2016:TMM,Liu2014}, abstract paintings \cite{alameda2016} or music \cite{Yazdani2011}. Although it is possible for the perceived emotion to differ from the intended expression in the video, like jokes falling flat, we find such cases to be uncommon in the datasets we used. 

We identify three major challenges faced by video emotion understanding. First, usually only a small subset of video frames directly depicts emotions, whereas other frames provide context that is necessary for understanding the emotions. Thus, the recognition method must be sufficiently sensitive to sparse emotional content. Second, there is usually one dominant emotion for every video, but other emotions could make interspersed appearances. Therefore, it is important to distinguish the video segments that contribute the most to the video\rq{}s overall emotion, a problem known as video emotion attribution \cite{heterog_tac}. Third, in comparison to commercial production, user-generated videos
are highly variable in production quality and contain diverse objects, scenes, and events, which hinders computational understanding. 

Observing these challenges, we argue that it is crucial to extract feature representations that are sensitive to emotions and invariant under conditions irrelevant to emotions from the videos. In previous work, this is achieved by combining low-level and middle-level features \cite{baohan2014AAAI}, or by using auxiliary image sentiment dataset to encode video frames \cite{heterog_tac,baohan_icmr}. The effectiveness of these features has been demonstrated on three emotion-related vision tasks, including emotion recognition, emotion attribution, and emotion-oriented video summarization. However, a major drawback of previous work is the three tasks were tackled separately and cannot inform each other.

Extending our earlier work \cite{emotion_net2017}, we propose a multi-task neural architecture, the Bi-stream Emotion Attribution-Classification Network (BEAC-Net), which tackles both emotion attribution and classification at the same time, thereby allowing related tasks to reinforce each other. BEAC-Net is composed of an attribution network (A-Net) and a classification network (C-Net). The attribution network learns to select a segment from the entire video that captures the main emotion. The classification network processes the segment selected by the A-Net as well as the entire video in a bi-stream architecture in order to recognize the overall emotion. In this setup, both the content information and the emotional information are retained to achieve high accuracy with a small number of convolutional layers. Empirical evaluation on the Ekman-6 and the Emotion6 Video datasets demonstrate clear benefits of the joint approach and the complementary nature of the two streams. 

The contributions of this work can be summarized as follows: (1) We propose BEAC-Net, an end-to-end trainable neural architecture that tackles emotion attribution and classification simultaneously with significant performance improvements. (2) We propose an efficient dynamic programming method for video summarization based on the output of A-Net. (3) To establish a good benchmark for emotion attribution, we re-annotate
the Ekman-6 dataset with the most emotion-oriented segments which can be used as
the ground-truth for the emotion attribution task. 

\section{Background and Related Work}

\begin{figure*}[t] \begin{centering} 
\includegraphics[scale=0.5]{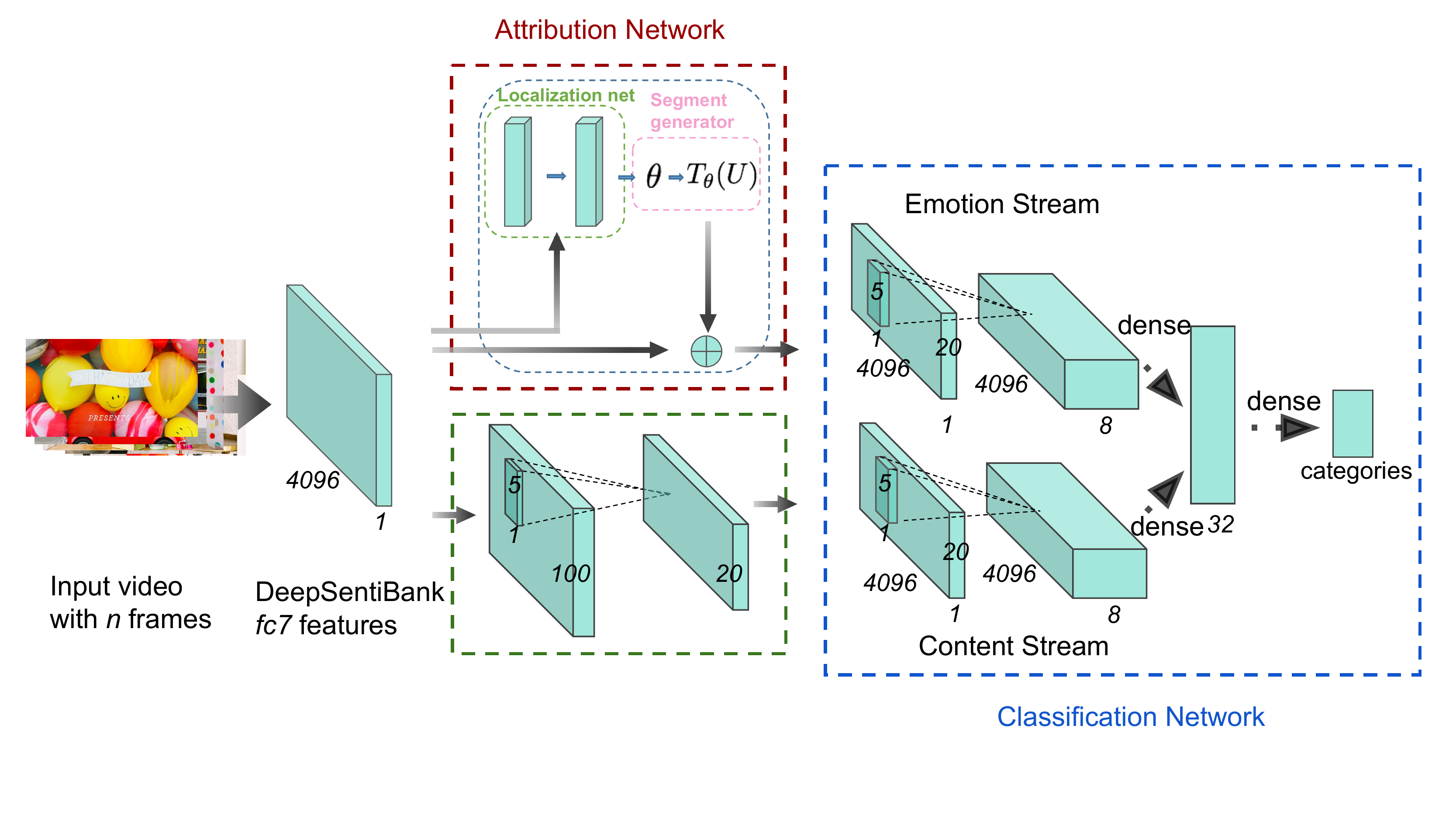}
\par
\end{centering} 
\caption{\label{fig:Overview} An overview of the BEAC-Net neural architecture.
We extract features from every video frame from the ``fc7'' layer of the DeepSentiBank convolutional neural network model. The attribution network extracts one video segment that expresses the dominant emotion, which is fed to the emotion stream of the classification network. The whole video is downsampled using convolution along the temporal dimension (dashed green box) and fed to the content stream of the classification network. }
\end{figure*}

\subsection{Psychological Theories and Implications}

\noindent Most works on recognizing emotions from visual content follow psychological theories that lay out a fixed number of emotion categories, such as Ekman's six pan-cultural basic emotions \cite{Ekman1972,Ekman1999} and Plutchik's wheel of emotion \cite{plutchik1980emotion}. These emotions are considered to be ``basic'' because they are associated with prototypical facial expressions, verbal and non-verbal language, distinct antecedent events, and physiological responses. The emotions constantly affect our expression and perception via appraisal-coping cycles throughout our daily activities \cite{Gross2002}, including video production and consumption.

Recent psychological theories \cite{Barrett2006,Lindquist2013} suggest the range of emotions is far more varied than prescribed by basic emotion theories. The psychological constructionist view argues that emotions emerge from other, more basic cognitive and affective ingredients. For example, bodily sensation patterns can be identified for both basic and non-basic emotions \cite{Nummenmaa2013}. The ingredients that form emotions include interaction among cognitive processes, temporal succession of appraisals, and coping behaviors \cite{Li:Humor2015,Gross2002}. This may have inspired computational work like DeepSentiBank \cite{TaoChen2014Deepsentibank} and zero-shot emotion recognition \cite{heterog_tac}, which broaden the emotion categories that can be recognized. 

In a similar vein, dimensional theories of emotion \cite{dimenaffects1980,fontaine2007world,lovheim2012new} characterize emotions as points in a multi-dimensional space. These theories also allow richer emotion descriptions than the basic categories. Early works almost exclusively use the two dimensions of valence and arousal \cite{dimenaffects1980}, whereas more recent theories have proposed three \cite{lovheim2012new} or four dimensions \cite{fontaine2007world}. 
To date, most computational approaches that adopt the dimensional view \cite{AVEC2017challenge1,AVEC2017challenge2,baveye2015liris} employ valence and arousal. Notably, \cite{Benini2011} proposes a three-dimensional model for movie recommendation, where the dimensions include passionate vs. reflective, fast vs. slow paced and high vs. low energy. 

Though we recognize the validity of recent theoretical developments, in this paper we adopt the six basic emotion categories for practical reasons, as these categories provide a time-tested scheme for classification and data annotation. 

\subsection{Multimodal Emotion Recognition}

Researchers explored features for visual emotion recognition, such as features enlightened by psychological and art
theories \cite{Machajdik2010} and shape features \cite{Lu2012}.
Jou et al. \cite{Jou2014} focused on animated GIF files, which are similar to short video clips. Sparse coding\cite{MI-SC,Song2013} also proves to be effective for emotion recognition. 

Facial
expressions have been used as a main source of information for emotion recognition \cite{Joho2009, Zhao2011}. Zhen et al. \cite{Zhen2016:TMM} create features by localizing facial muscular regions. Liu et al. \cite{Liu2014} construct
expressionlet, a mid-level representation for dynamic facial expression
recognition. 

Combining multimodal information with visual input is another promising direction. 
A number of works recognize emotions and/or affects from speech \cite{schuller2003hidden,mao2014learning,Zhang2018}.
Wang et al \cite{Wang2006} adapted audio-visual features to classify 2040 frames of 36 Hollywood
movies into 7 emotions. 
\cite{zeng2007audio} jointly uses speech and facial expressions. \cite{acar2016comprehensive} extracts mid-level audio-visual features.  
 \cite{DeepMultimodalLearning} employs the visual, auditory, and textual modalities for video retrieval. \cite{kahou2013combining} provides a comprehensive technique that exploits audio, facial expressions, spatial-temporal information, and mouth movements. 

  Deep neural networks have also been used for visual sentiment analysis
\cite{You2015AAAI_img_sentiment,Borth2013acmmm}. A large-scale visual
sentiment dataset was proposed in Sentibank \cite{Borth2013acmmm} and
DeepSentiBank \cite{TaoChen2014Deepsentibank}. Sentibank is composed of 1,533
adjective-noun pairs, such as ``happy dog'' and
``beautiful sky''. Subsequently, the authors used deep convolutional neural networks (CNN) to deal with images of strong sentiment and achieved improved performance.  For a recent survey on understanding emotions and affects from video, we refer readers to \cite{Wang-Ji2015}.

Most existing works on video emotion understanding focus on classification. As emotional content is sparsely expressed in user-generated videos, the task of identifying emotional segments in the video \cite{Arifin2008,baohan_icmr,heterog_tac} may facilitate the classification task. 
Noting the synergy between the two tasks, in this paper, we propose a multi-task neural network that tackles both simultaneously. 

\subsection{Emotion-oriented video summarization}

\noindent Video summarization has been studied for more than two decades
\cite{Truong:2007:VAS:1198302.1198305} and a detailed review is beyond the
scope of this paper. In broad strokes, we can categorize summarization techniques into two major classes: keyframes extraction
and video skims. A large variety of video features have been
exploited, including visual saliency\cite{Ma:2002:UAM:641007.641116}, motion cues
\cite{lai2012key}, mid-level features \cite{event_driven_summary,WangNgo2012}, and semantic
recognition \cite{DBLP:conf/mm/WangJCGDW14}.

Inspired by the task of semantic attribution in
text analysis, the task of emotion attribution \cite{heterog_tac} is defined as
attributing the video's overall emotion to its individual segments. In
\cite{heterog_tac}, emotion recognition, summarization and
attribution tasks are tackled individually. Intuitively, emotion
recognition can benefit from emotion attribution, which identifies emotional segments. Based on this insight, we propose to solve the two tasks together in an end-to-end manner. 

In the previous work \cite{emotion_net2017}, we focused on only the emotional segment and neglected other frames, which may serve as the context for understanding emotional activities. In this paper, the emotion segment and its context are combined within a two-stream architecture.

\subsection{Spatial-temporal Neural Networks}

\noindent The proposed technique in this paper is inspired partially by the Spatial Transformer Network (ST-Net) \cite{jaderberg2015spatial}, which is firstly proposed for image classification. ST-Net
provides the capability for fully-differentiable spatial transformation, which facilitates tasks like co-localization
\cite{singh2016end} and spatial attention \cite{show2015tell}. 

ST-Net could be split into three parts: 1) Localization Network, which outputs parameters $\bm \theta$ that control a spatial transformation applied on the input feature map. 2) Parameterized Sampling Grid, which realizes the spatial transformation by mapping individual elements on the input feature to elements on a regular grid $G$. This mechanism supports a broad list of transformations, such as rotation, reflection, cropping, scaling, affine transformation, etc. 3) Differentiable Image Sampling, a sampling kernel that interpolates the input feature map to handle fractional coordinates. This mechanism is differentiable and hence allows gradient information from later stages to propagate back to the input.  

There are several variants of ST-Net. Singh and Lee\cite{singh2016end} proposed a loss function to solve the problem that ST-Net's
output patch may extend beyond the input boundaries. Lin et al.\cite{Lin2017} improved upon
ST-Net by connecting it to the inverse compositional LK algorithm.

The attribution network in BEAC-Net can be seen as performing a similar transformation on the temporal dimension. It enables the network to identify video segments that carry emotional content, which alleviates the sparsity of emotion content in videos.

BEAC-Net contains a two-stream architecture that extracts features not only from the video segment identified by the attribution network but also the entire video as its context. This is different from the two-stream architecture introduced by \cite{simonyan2014two}, which contains a convolutional stream to process pixels of the frames and another convolutional stream for optical flow features.  \cite{carreira2017quo} provides a further generalization to 3D convolutions. By leveraging local motion information from optical flow, these approaches are effective at activity recognition. Optical flow features are not used in this paper, though we believe they could lead to further improvements.

\section{The Emotion Attribution-Classification Network }
BEAC-Net is an end-to-end multi-task network that tackles both emotion recognition and attribution. 
In this section, we describe its two constituents: the emotion attribution network (A-Net) and the emotion classification network (C-Net). The former extracts a segment from the video that contains emotional content, whereas the latter classifies the video into an emotion by using the extracted segment together with its context.  
Each input video is represented by features extracted using the CNN trained on DeepSentiBank, as described in \cite{TaoChen2014Deepsentibank}. Fig. \ref{fig:Overview} provides an illustration of the network architecture. 

\subsection{Feature Extraction}
We extract video features using the deep convolutional network provided by \cite{TaoChen2014Deepsentibank}, which classifies images into adjective-noun pairs (ANPs). Each ANP is a concept consists of an adjective followed by a noun, such as \lq{}\lq{}creepy house\rq{}\rq{} and \lq{}\lq{}dark places\rq{}\rq{}. The network was trained on $867,919$ images for classification into $2,089$ ANPs. 
The network in \cite{TaoChen2014Deepsentibank} contains five convolutional layers and three fully connected layers. We take the 4096-dimensional activation from the second fully connected layer labeled as ``fc7''. 
The classification into ANPs can be considered as the joint recognition of objects and the emotions associated with the object. We believe the features extracted by this network retain both object and affective information from the images.

Formally, let us denote the whole dataset as $\mathcal{D}=\{(X_i, y_i, \bm \alpha_i)\}_{i=1,\ldots,N}$ where $X_i$ denotes the $i^\text{th}$ video , $y_i$ denotes its emotion label, and $\bm \alpha_i$ denotes the supervision on emotion attribution, which is explained in the next section. The $M$ frames of $X_i$ are denoted as $\bm x_i = \{\bm x_{i,t}\}_{t=1,\ldots,M}$. Let $\phi(\cdot)$ be the feature extraction function; the $i^\text{th}$ video features are represented as $\phi(\bm x_{i})$.

\subsection{The Attribution Network}
\label{subsec:Localization-Network}
The emotion attribution task is to identify video frames responsible for a particular emotion in the video. 
The attribution network learns to select one continuous segment of $L$ frames that contains the main emotion in the original video of $M$ frames. The network predicts two parameters $\alpha_1$ and $\alpha_2$, which are in the fixed range [-1,1] and sufficient for selecting of any continuous video segment. This formulation simplifies training and inference.  

Formally, the indices of frames are in the range $[1, M]$. We let the indices be continuous due to the possibility of interpolation. For any given starting frame $t_s$ and ending frame $t_e$ ($t_e > t_s$), we can compute $\bm \alpha = (\alpha_1 \; \alpha_2)$ as follows. 
\begin{equation} 
\begin{split}
\alpha_{1} & =\frac{1}{M}(t_e-t_s) \\
\alpha_{2} & =\frac{1}{M}(t_e+t_s)-1
\end{split}
\label{eq:trans} 
\end{equation}
Obviously, $\alpha_1 \in (0,1]$ and $\alpha_2 \in [-1,1]$. 
We then define the transformation function $f_{\bm \alpha}(t): [t_s, t_e] \mapsto [-1, 1]$ as:
\begin{equation}  
f_{\bm \alpha}(t) = \frac{2t}{\alpha_1 M}  - \alpha_2 - 1
\label{eq:transformer} 
\end{equation}

%


Therefore, $\bm \alpha$ sufficiently parameterizes the frame-selection operation. We use two fully connected layers to project the video features $\phi(\bm x_{i})$ to $\bm \alpha$. In order to solve the emotion attribution task, we perform the inverse operation of Eq. \ref{eq:trans} and recover the start time $t_s$ and end time $t_e$ from the regression output $\hat{\bm \alpha}$:
\begin{equation}
\begin{split}
\hat{t}_s& =\frac{M}{2}(\hat{\alpha}_{2}-\hat{\alpha}_{1}+1)\\
\hat{t}_e& =\frac{M}{2}(\hat{\alpha}_{1}+\hat{\alpha}_{2}+1)
\end{split}
\label{eq:inverse_tran}
\end{equation}

The network utilizes the following square loss function for regression, which computes the differences between the output of the network $\hat{\bm \alpha}_i$ and the externally supplied supervision $\bm \alpha_i$.
\begin{equation} \mathcal{L}^{A}_i = \left(  \left(\alpha_{i, 1}-\hat{\alpha}_{i, 1}\right)^{2}+\left(\alpha_{i, 2}-\hat{\alpha}_{i, 2}\right)^{2} \right) \label{eq:STloss} 
\end{equation} 
For training, we interpolate video features when $\hat{t}_s$ and $\hat{t}_e$ are not integers. During inference, they are rounded to the nearest integer.

\subsection{The Classification Network}
\label{subsec:The-Proposed-Framework}
The emotion recognition task, as the name implies, classifies a video as one of the emotions. We propose a novel two-stream neural architecture that employs the emotion segment selected by the attribution network in combination with the original video. 
This architecture allows us to focus on the dominant emotion, but also consider the context it appears in. It may also be seen as a multi-resolution network, which applies coarse resolution on the entire video and fine resolution on the emotional segment. 
 We call the two streams the emotion stream and the content stream respectively. 
 
Before the content stream, we perform a temporal convolution 
in order to compress $M$ frames from the video to $L$ frames. This results in an input dimension $L\times4096$, identical to the input to the emotion stream. 

The two streams have symmetrical architectures, containing one convolution layer before two fully connected layers, followed by a final, softmax function. The convolution parameters are shared between the streams, which we find to accelerate training. It is worth noting that the two streams capture complementary information and the interaction between streams is  critical for high accuracy (see the ablation experiments in Section \ref{section:recognition_results}). 

The classification network adopts the standard cross-entropy loss. For a $K$-class classification, the loss function is written as
\begin{equation} \mathcal{L}^{C}_i=\sum_{k=1}^{K}-y_{ik}\cdot \log(\hat{y}_{ik})\label{eq:ECloss} \end{equation}
where $y_{i}$ is a ground-truth one-hot vector and $\hat{y}_{i}$ is the output of the softmax function. 

\subsection{Joint Training}

In order to stabilize optimization, for every data point $X_i$, we introduce an indicator function $\mathds{1}(o_i \ge \beta)$, where $o_i$ indicates the temporal intersection over union (tIoU) between the A-Net\rq{}s prediction $\hat{\bm \alpha}_i$ and ground truth $\bm \alpha_i$. $\beta$ is a predefined threshold. That is, the indicator function returns 1 if and only if the attribution network is sufficiently accurate for $X_i$. 

We combine the standard cross-entropy classification loss and the attribution regression loss to create the final loss function as
\begin{equation} \mathcal{L} = \frac{1}{N}\sum_{i}\left[ \mathds{1}(o_i\ge\beta) \mathcal{L}_{i}^C + \mathds{1}(o_i<\beta) \mathcal{L}_{i}^A\right] \label{eq:totalloss} 
\end{equation}
In plain words, although A-net and C-net are trained jointly, gradients from the classification loss are backpropagated only when the attribution network is accurate enough. Otherwise, only the gradients from the attribution loss are propagated backward and the parameters from C-net remain the same. For each data point, the network focuses on training either the A-Net or the C-Net, but not both. We find this to stabilize training and improve performance. 

\subsection{Emotion-Oriented Summarization}
In this section, based on the output of the emotion attribution network, we formulate the emotion-oriented summarization problem as a constrained optimization problem. The summarization aims to maintain continuity between selected video frames while select as few frames as possible and focus on the emotional content. This problem can be efficiently solved by MINMAX dynamic programming \cite{minmax}. 

The emotion-oriented summarization problem can be formally stated as follows. From a video $X_i$ containing $M$ frames $\{\bm x_{i,j}\}_{j=1,\ldots,M}$, we want to select a subset of frames $h_1, \ldots, h_P\in \{1, \ldots, M\}$ that minimizes the sum of individual frame\rq{}s cost: 
\begin{equation}
\min \sum_p \text{cost}(h_p)
\end{equation}
subject to the following constraints. 
\begin{itemize}
\item $h_1 = 1, h_P = M$. Always select the first and the last frames.
\item $h_{p+1} - h_{p} \le K_{max}$. The frames are not too spaced out. The constant $K_{max}$ is the maximum index difference for adjacent summary frames. 
\item $\forall h_p \le i, j \le h_{p+1}, d(\phi(\bm x_{i}), \phi(\bm x_{j})) \le D_{max}$, where $d(\cdot)$ is the Euclidean distance. In words, there is no large feature-space discontinuity ($\le D_{max}$) between $h_{p}$ and $h_{p+1}$ in the video. 
\end{itemize}
In other words, we minimize the total cost by selecting fewer frames, but we must also make sure removing a frame does not create a large gap in feature space. This prevents large discontinuity from disrupting the viewing experience.  
 
Based on the emotional segment identified by the A-Net, we encourage the inclusion of emotional frames in the summary by setting 
\[ \text{cost}(i)=\begin{cases} 1 &
\text{if} \; \bm x_i \; \text{in the emotional segment}\\ 2 &
\text{if} \; \bm x_i \; \text{not in the emotional segment} \end{cases} 
\]

We present a solution to the problem using the MINMAX dynamic programming technique \cite{minmax}.    
We measure the discontinuity between adjacent frames in the video summary between the selected frames $h_{p}$ and $h_{p+1}$ as

\[ D_{p}^{p+1}=
\begin{cases}
\underset{i,j\in\left[h_p,h_{p+1}\right]}{\max}d\left(\phi(\bm x_{h_i}), \phi(\bm x_{h_j})\right) & if\:h_{p+1}-h_{p}\leq K_{max}\\ \infty & otherwise
\end{cases} 
\]
where $d(\cdot)$ denotes the Euclidean distance between the features $\phi(\bm x_{h_i})$ and $\phi(\bm x_{h_j})$. The cost of this segment, denoted by $R_{p}^{p+1}$, is

\[ R_{p}^{p+1}=
\begin{cases} \text{cost}(h_p) &
\text{if} \:D^{p+1}_p \leq D_{max}\\ \infty & otherwise \end{cases} \]

\noindent This requires the discontinuity in every segment to be smaller than the
maximum allowable amount $D_{max}$. If the sequence
segment has an admissible discontinuity, the cost of the segment is represented by the cost of the summary frame. 

Using the dynamic programming technique, we define the quantity $N\left(t,h_{p+1}\right)$ as the minimum cost where $t$ is the number of frames selected so far and $h_{p+1}$ is the next frame to select. The recurrence equation is given by
\begin{equation}
N\left(t,h_{p+1}\right)=\underset{h_p \in [h_{p+1}-K_{max},h_{p+1}]}{\mathrm{min}} \left(
N\left(t-1,h_p\right)+R^{p+1}_{p}\right)
\end{equation}
for all $R_{p}^{p+1}<\infty$.
To obtain the optimal solution, we find the minimum value $\min_t N(t, M)$ because we must include the last frame of the original value. The whole sequence is found by tracing through the intermediate minima back to the first frame. It is easy to see that the time complexity of the algorithm is $O(MK_{max}T_{max})$, where $T_{max}$ is the maximum number of frames that the video summary can have. 

\section{Experiments}

\subsection{Dataset and Preprocessing}

We conduct experiments on two video emotion datasets based on Ekman's six basic emotions.

\vspace{0.05in}

\noindent \textbf{The Emotion6 Video Dataset}. The Emotion6 dataset \cite{emotion6} 
contains 1980 images that are labeled with a distribution over 6 basic emotions (anger, surprise, fear, joy, disgust, and sadness) and a neutral category. The images do not contain facial expressions or text directly associated with emotions. We consider the emotion category with the highest probability as the dominant emotion. 

For the purpose of video understanding, we create Emotion6 Video (Emotion6V), a synthetic dataset of emotional videos using images from Emotion6. We collected an auxiliary set of neutral images from the first few seconds and the last few seconds of YouTube videos as these frames are unlikely to contain emotions. After the frames are collected, we manually examine these frames and select a subset that contains no emotions. 

In order to create a video with a particular dominant emotion, we select images from Emotion6 that have the dominant emotion or from the neutral set. This allows us to create ground-truth emotion labels and duration annotations for the emotional segment. We created 600 videos for each class for a total of 3,600 videos. 

\vspace{0.05in}

\noindent \textbf{The Ekman-6 Dataset.} The Ekman-6 dataset 
\cite{heterog_tac} contains 6 basic types of emotions: anger, surprise,
fear, joy, disgust, and sadness. The total number of videos is 1637. In this paper, we use 1496 videos whose sizes are greater than 45MiB, which are composed of 213 videos labeled as anger, 331 as surprise, 276 as fear, 289 as joy, 195 as sadness and 192 as disgust.
To further assist the tasks of attribution and video-oriented summarization, every video is annotated with the most significant emotion segment. For every video, three annotators selected no more than 3 key segments that contribute the most to the overall emotion of the video. The longest overlap between any two annotators was considered to be the ground truth.

\vspace{0.05in}

\noindent \textbf{Preprocessing.} We use the same split for the two datasets, with 70\% of the data used as the training set, 15\% as validation, and 15\% for testing. As a preprocessing step, we
uniformly sample $30$ frames for each video in the Emotion6 Video dataset. Due to the fact that videos in the Ekman-6 dataset are slightly longer than Emotion6V, we uniformly sample $100$ frames from each video. Black frames are added if the video contains less than $100$ frames. Less than 1\% of the videos comprise less than 100 frames, so the padding is rarely necessary. We also create two variations for the Ekman-6 dataset. The two-class condition focuses on the two largest emotion categories, anger and surprise. The second condition employs all videos in the dataset. The data and source code are available at \url{https://github.com/guoyuntu/BEAC_network}.

\subsection{Hyperparameter Settings}

The convolution layer before the content stream has a single filter with kernel size $5 \times 1$ (temporal dimension $\times$ feature dimension) and stride $5 \times 1$.
The convolutional layer in the classification network, which is shared by both streams, has 8 convolutional filters with $5\times1$ kernels and $1 \times 1$ stride. The two fully connected layers have $32$ units each. The threshold $\beta$ in the loss function is set to 
0.6. The parameter $M$ is the overall video length and is set to 30 and 100 for Emotion6V and Ekman-6, respectively. The extracted length $L$ is always set to 20.  

We set the initial value of
$\alpha_1$ and $\alpha_2$ to $0.5$ and $0$ respectively. The models are trained for 200 epochs. Dropout is
employed here on all fully-connected layers and the keep ratio is set as $0.75$. The network is optimized using Adam\cite{kingma2014adam}. For each
dataset, experiments are repeatedly 5 times
and the averaged performance is reported. 


\begin{table*}[th]
\caption{\label{tab:video-classification} Emotion recognition results. }
\renewcommand{\arraystretch}{1.3}
\centering
\begin{tabular}{llllp{1.4cm}lp{1.4cm}p{1.4cm}l} \toprule
Dataset  & SVM  & ITE\cite{heterog_tac}  & C-Stream  & Unsup. E-Stream & E-stream & C+UnsupE & Temporal Attention & BEAC-Net 
\\
\midrule
Emotion6 Video  &  $80.0\%$  & $77.5\%$  & $81.3\%$  & $82.2\%$  & $99.5\%$ & $88.9\%$ & $91.0\%$ & $\mathbf{99.7\%}$
\\
Ekman-6 (two classes) & $62.8\%$  & $65.3\%$  & $59.5\%$  & $68.9\%$  & $70.4\%$ & $69.8\%$ & $60.5\%$ & $\mathbf{71.6\%}$
\\
Ekman-6 (all classes)  & $42.8\%$  & $43.2\%$  & $47.1\%$  & $41.7\%$  & $44.9\%$ & $47.1\%$ & $39.9\%$ & $\mathbf{49.3\%}$
\\ \bottomrule
\end{tabular}
\end{table*}

\subsection{Competing Baselines}

Our model is compared to the following baseline models.

\vspace{0.05in}

\noindent \textbf{Image Transfer Encoding (ITE).} Our model is compared against the
state-of-the-art method \textendash{} Image Transfer Encoding (ITE) \cite{heterog_tac}, which uses an emotion-centric
dictionary extracted from auxiliary images to encode videos. The encoding scheme has been shown to have a good performance in emotion recognition, attribution, and emotion-oriented summarization. 

We replicated the same setting of ITE as described in \cite{heterog_tac}: we first cluster a set of  auxiliary images into 2000 clusters using K-means on features extracted from AlexNet\cite{AlexNet2012}. For each frame, we select $K$ clusters whose center are the closest to the frame and the video feature vector is computed as the sum of the individual frames' similarity to the $K$ clusters. Formally, let $\{\bm c_d\}_{d=1, \ldots, 2000}$ denote the cluster centers. The representation for the $i^\text{th}$ video is a 2000-dimensional vector $\bm s_i$, which is computed as a summation over all frames:
\begin{equation}
s_{i,d}=\sum_{j=1}^M \cos(\phi(x_{i,j}), c_d)\; \mathbbm{1}\left(c_d \in \text{KNN}(x_{i,j})\right)
\end{equation}
where the indicator function equals 1 if and only if the cluster center $c_d$ is among the $K$ nearest clusters of $x_{i,j}$. 
A linear SVM model is trained for the emotion recognition task. The
attribution can be solved by selecting a sequence of frames whose similarities to video-level representation are greater than a predefined threshold while no more than 10 consecutive frames falling below the threshold. The frames with maximal similarities to the video-level representation are chosen as the summary of the video.

\vspace{0.05in}

\noindent \textbf{Support Vector Machine (SVM).} The $fc7$ features are used to train a linear SVM classifier on each frame of the video. The majority vote is extracted as the final classification label. The attribution results are obtained by selecting the longest segment classified as the same emotion and the summarization results are obtained by selecting the frames with the highest emotion scores. 

\vspace{0.05in}

\noindent \textbf{The Content Stream Only (C-Stream).} For the task of video emotion classification, we perform an ablation study by removing the attribution network and the associated emotion stream from the classification network. What remains is a single-stream, conventional convolutional neural network. We report the result for emotion classification only, as this network is not capable of emotion attribution.

\vspace{0.05in}

\noindent \textbf{Supervised Emotion Stream (E-Stream).} As a second ablated network, we remove the content stream from the classification network. The attribution network and the associated emotion stream are kept intact. The attribution loss is also kept as part of the loss function. 

\vspace{0.05in}

\noindent \textbf{Unsupervised Emotion Stream (Unsup. E-Stream).} This is a third ablated network. Similar to the E-Stream version, we remove the content stream from the classification network. In addition, we also remove the attribution loss from the loss function. The A-Net and the emotion stream are kept intact. 
That is, we use only the emotion stream for classification, but do not supply supervision to the attribution network. 

\vspace{0.05in}

\noindent \textbf{C-Stream and Unsupervised E-Stream (C+UnsupE).} This is a fourth ablated network. We use both the C-stream with E-Stream but remove the attribution loss. This is equivalent to the full BEAC-Net \emph{sans} the supervision signal for emotion attribution. 

\vspace{0.05in}

\noindent \textbf{Temporal Attention.} Due to the popularity of the attention mechanism (e.g., \cite{Anderson2018}) in neural networks, we create a baseline using a typical attention formulation over the temporal dimension (e.g., \cite{LiYao2015}). We modify the A-Net by adding two fully connected layers with 128 hidden units and the ReLU activation function, followed by a softmax operation. The output is an attention weight $\beta_{i,t}$ for every frame $t$ in the $i^{\text{th}}$ video, such that $\beta_{i,t} > 0$ and $\sum_t \beta_{i,t} = 1$. Recall that the features of frame $t$ in the $i^{\text{th}}$ video are denoted as $\phi(\bm x_{i,t})$. The final representation for the entire video is computed as the convex combination $\sum_t \beta_{i,t} \phi(\bm x_{i,t})$ and fed to the E-stream. The E-stream and C-Stream remain unchanged from the full BEAC-Net.

\subsection{Results and Discussion}
\label{section:recognition_results}
\noindent \textbf{Emotion recognition}. We perform emotion recognition on the Emotion6 Video dataset and the Ekman-6 dataset, where Ekman-6 has two experimental conditions with different numbers of classes. Table 1 reports the classification accuracy, which is the simple average across different emotion categories. 

We observe that BEAC-Net achieves the best performance among all baseline models, including all ablated versions. Compared to the previous state-of-the-art method ITE, BEAC-Net improves classification performance by 22.2\%, 6.3\% and 6.1\%, respectively. 

The three experimental conditions establish an easy-to-hard spectrum. The artificial Emotion6 Video dataset is the simplest, for which a simple SVM can achieve 80\% accuracy. The full Ekman-6 with all 6 emotions is the most difficult. It is worth noting that the effectiveness of the bi-stream architecture is the most obvious on the most difficult full Ekman-6 dataset, leaving a 2.2\% gap between BEAC-Net and the second-best technique. E-Stream is almost the same as BEAC-Net on the simplest conditions, but the gap widens as the task gets more difficult. 

The ablation study reveals the complementarity of all constituents of BEAC-Net. The C-Stream convolutional network underperforms BEAC-Net by 18.4\%, 12.1\%, and 2.2\%. The E-Stream with attribution supervision underperforms by 0.2\%, 1.2\%, and 4.4\%. Interestingly, the E-Stream beats the C-Stream on the Emotion6V and two-class Ekman-6, but underperforms on the full Ekman-6 dataset. These results indicate that the two streams indeed complement each other under different conditions and their co-existence is crucial for accurate emotion recognition. 
The comparison between the unsupervised E-stream and E-stream as well as that between C+UnsupE and BEAC-Net demonstrate the benefit of the attribution supervisory signal. On average, the improvements on the three conditions are 14.1\%, 1.7\%, and 2.7\%, respectively. 

The comparison between Temporal Attention and C+UnsupE is particularly interesting due to their similarity. The only differences lie in the following. First, C+UnsupE uses hard cutoffs whereas the temporal attention baseline assigns a non-zero weight $\beta_t$ to every frame. Second, the A-Net selects a continuous video chunk, whereas the temporal attention may pay attention to arbitrary frames. Therefore, this comparison can help us understand if the proposed A-Net is better than the classical attention mechanism.

The results confirm the superiority of A-Net over temporal attention. Temporal attention performs better on the synthetic dataset, Emotion6V, by 2.1\%. However, C+UnsupE performs substantially better on the other two experimental conditions by margins of 9.3\% and 8.2\%, respectively. Since Ekman-6 is a natural dataset, we consider the performances on Ekman-6 to be more realistic and more representative. This result indicates that excluding many frames in the video is beneficial, corroborating our claim that A-Net's hard cutoff is effective in the handling of sparse emotional data.

\vspace{0.05in}
\noindent \textbf{Transferring Emotion Recognition.} In order to gain a better understanding of the two datasets, we perform an additional experiment on transfer learning. BEAC-Net is first trained on the full out-of-domain training set and finetuned for 10 epochs using only 20\% of the in-domain training set. This is compared to training BEAC-Net from scratch using the 20\% in-domain training set. 
Other settings, including learning rates, are kept identical to the emotion recognition experiments. 

The results, shown in Table \ref{tab:transfer-learning}, indicate that transfer learning in the low-data setting is advantageous. The benefit is the most pronounced when transferring from the real-world dataset, Ekman-6, to the synthetic dataset, Emotion6V, with an improvement of 61.8\%. This suggests learning on the real-world dataset creates transferrable representation of emotion information. 
Transferring from Emotion6V to Ekman-6 appears to be less beneficial, but still better than training from scratch. 
The synthetic construction of Emotion6 Video could be seen as a form of data augmentation. 

\begin{table}
\centering
\caption{\label{tab:transfer-learning} Transfer learning: out-of-domain BEAC-NET finetuned on 20\% in-domain data vs. training from scratch on 20\% in-domain.}
\begin{tabular}[t]{ccc}
\toprule
Test Set & Training Procedure & Accuracy\\
\midrule

\multirow{2}{*}{Emotion6 Video}& Ekman-6 $\rightarrow$ 20\% Emotion6V & \textbf{69.8}\% \\
& 20\% Emotion6V only & 18.0\% \\
& Chance & 16.7\% \\
\midrule
\multirow{2}{*}{Ekman-6} & Emotion6V $\rightarrow$ 20\% Ekman-6  &  \textbf{21.6}\% \\
  & 20\% Ekman-6 only & 20.7\% \\
  & Chance & 16.7\% \\
\bottomrule
\end{tabular}
\end{table}

\begin{figure*}[th] 
\centering{} \includegraphics[scale=0.36]{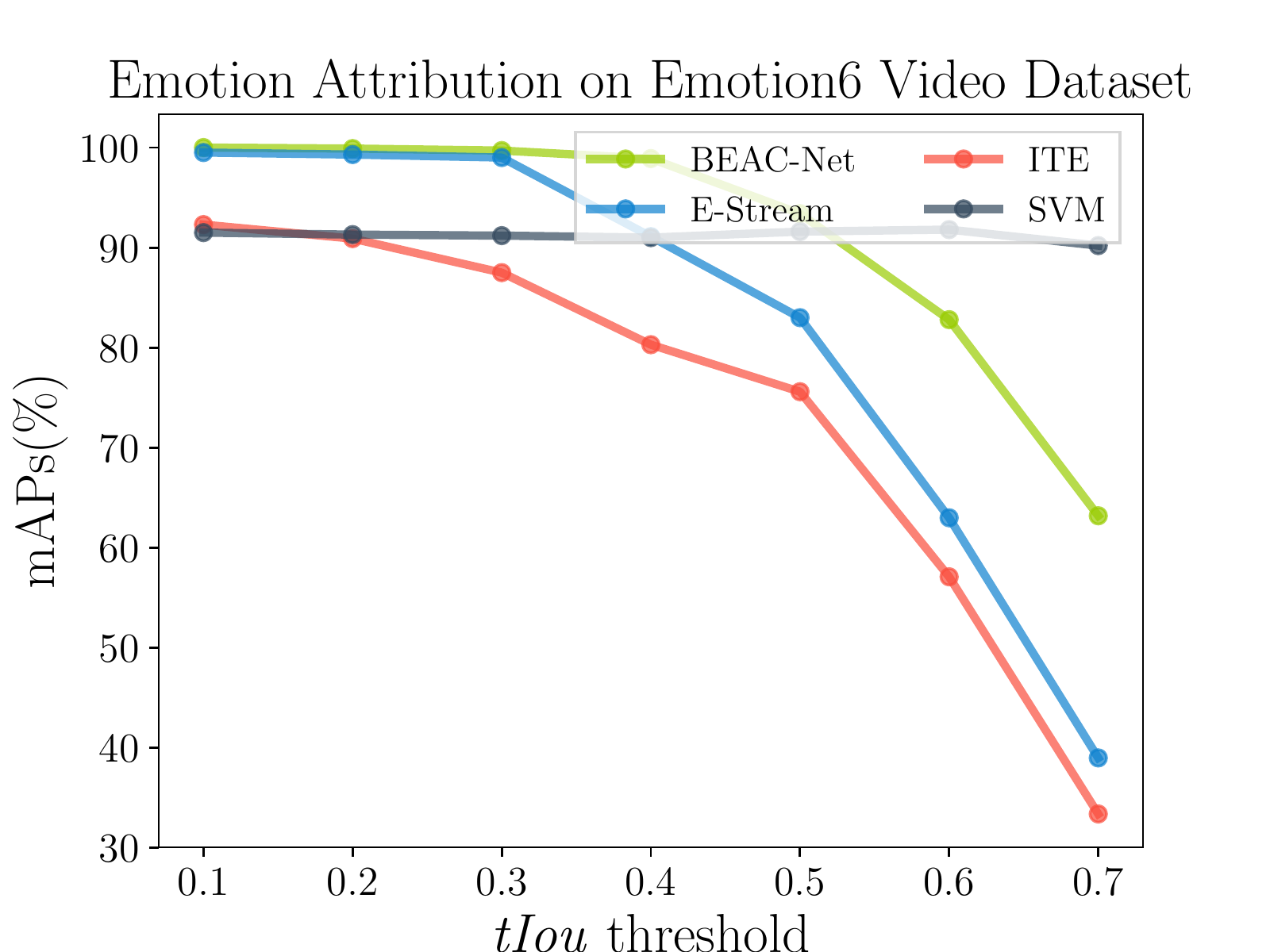}
\includegraphics[scale=0.36]{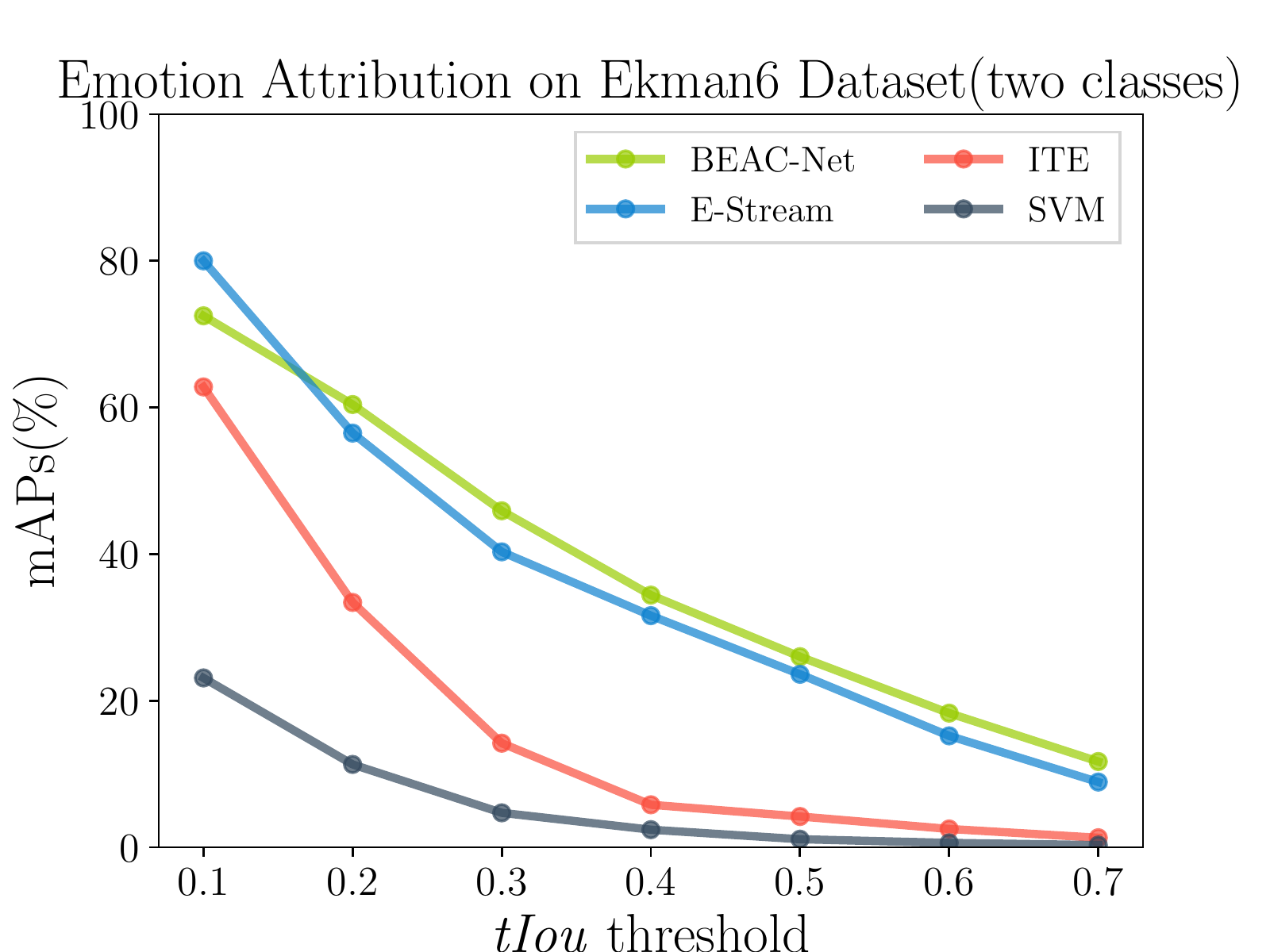}
\includegraphics[scale=0.36]{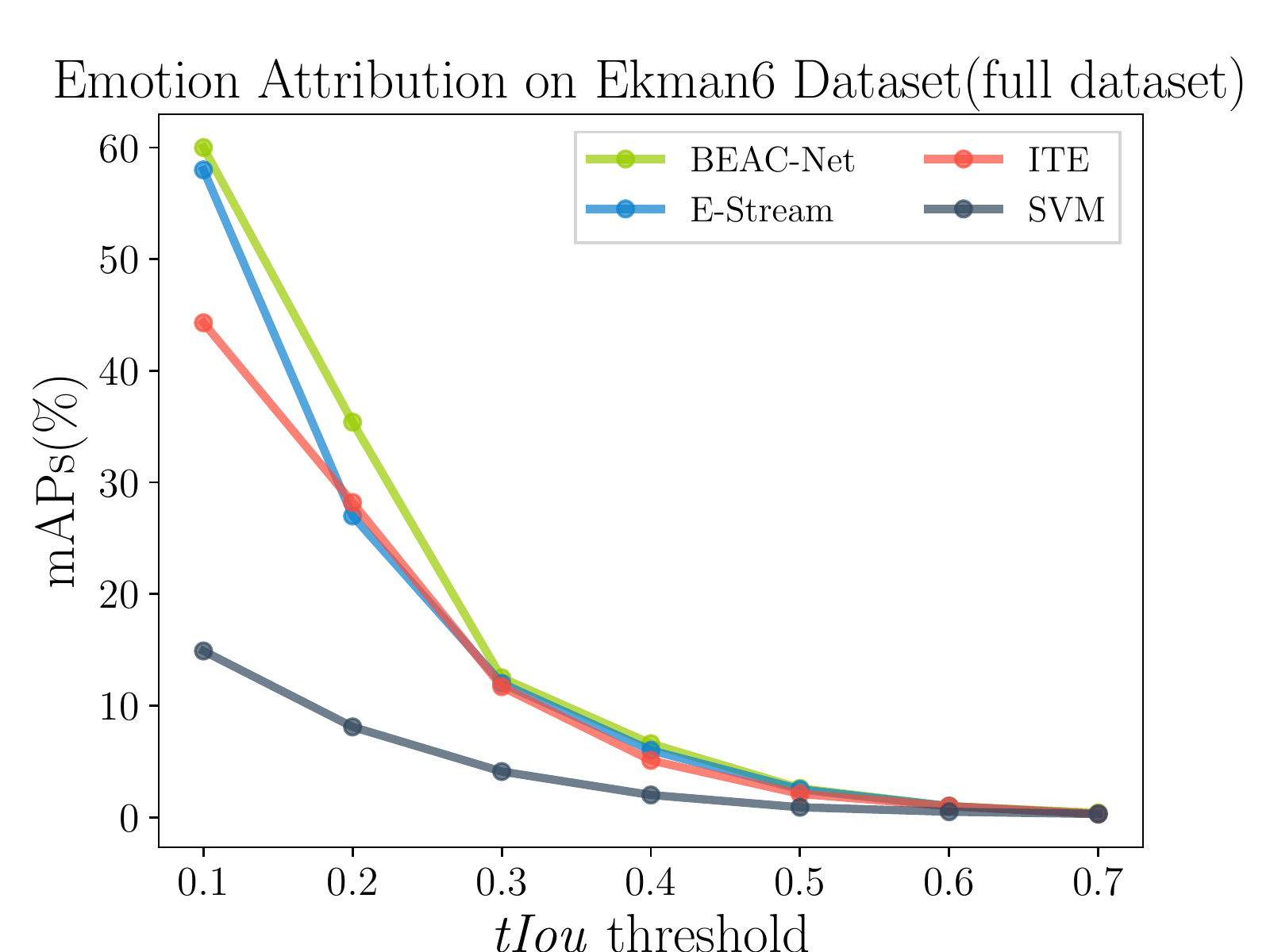} 
\caption{\label{tab:video-attribution-Ekman6} Emotion
attribution results. We report the mAP scores for each dataset. The horizontal axis indicates different tIoU thresholds}
\end{figure*}

\begin{figure}[th] \begin{centering} 
\includegraphics[scale=0.45]{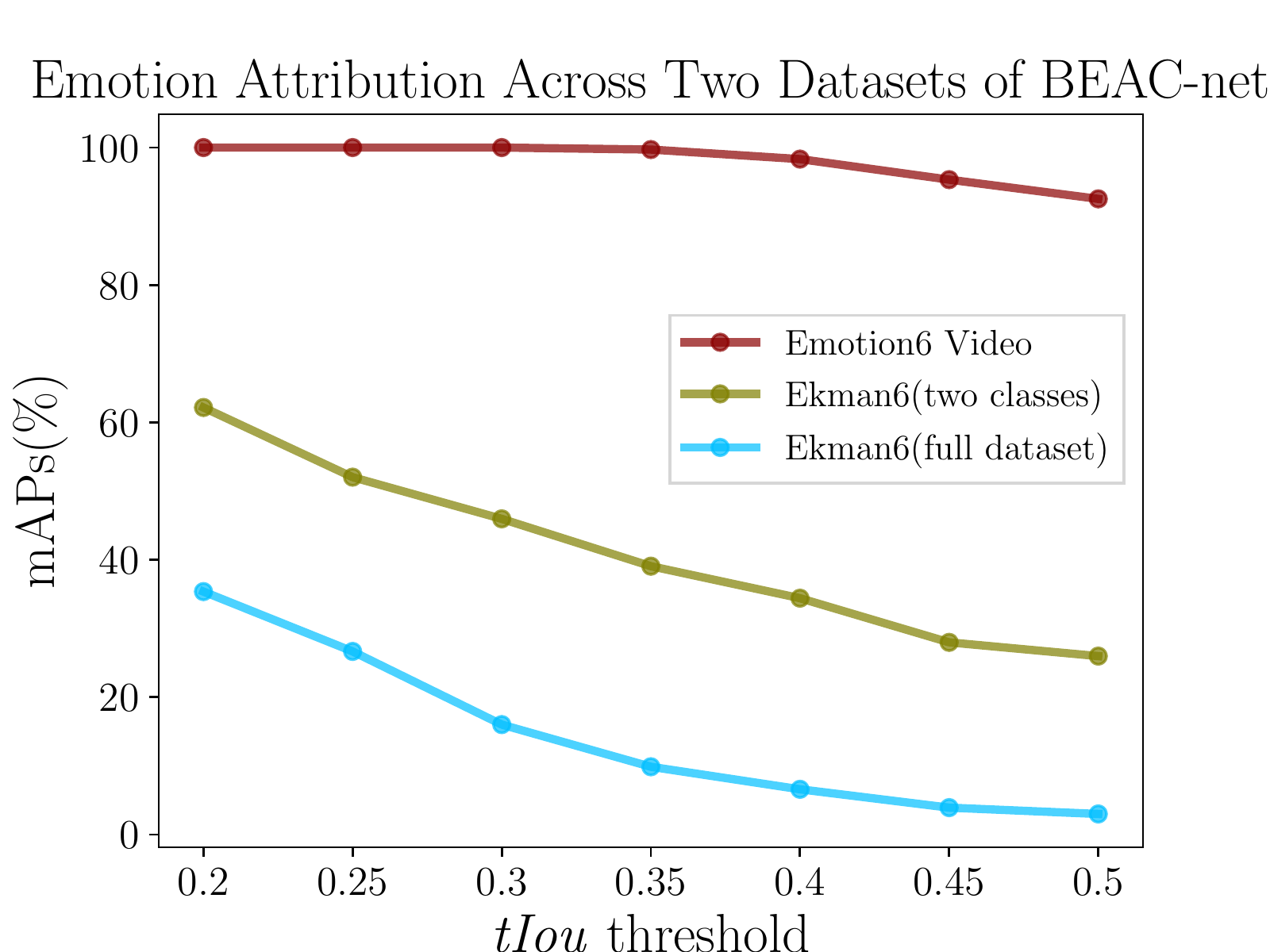}
\par
\end{centering} 
\caption{\label{fig:attri_across}  Attribution Result with same range of thresholds. Here we just report the result of BEAC-net.}
\end{figure}

\vspace{0.05in}

\noindent \textbf{Emotion Attribution.} 
We report the results on emotion attribution. Here the comparison baselines include 
ITE, SVM and Unsupervised E-Stream. For SVM, the longest majority-voted video emotion segment is considered to be the extracted emotional segment. 

We use mean average precision (mAP) to evaluate the performance of emotion
attribution, following the convention in video understanding \cite{caba2015activitynet}. To calculate the mAP, we first compute the overlap between the predicted video segment and the
ground-truth segment, as the temporal intersection over
union ($tIoU$). The predicted segment will be
marked as correct if the overlap is greater than a threshold, which varies from 0.1 to 0.7 in the experiments.
The three datasets, Emotion6 Video, two-class Ekman-6, and all-class Ekman-6 are the same as before. Fig.
\ref{tab:video-attribution-Ekman6} shows the results, where the horizontal axis indicate different tIoU thresholds.  

Once again, we observe strong performance from BEAC-Net, which achieves the best performance in almost all conditions. This validates that the A-Net can
help identify the video segments that contribute the most to the
overall emotion of one video. The unsupervised E-Stream performs worse than BEAC-Stream, but remains a close second in the Ekman-6 experiments. 

On the Emotion6V dataset, BEAC-Net outperforms other methods except for the last two tIoU thresholds, where the SVM method has very good and stable performance. We hypothesize that this is because every frame in Emotion6V was drawn from an image dataset, where every image is annotated with a definite emotion label. The SVM method classifies individual images to one label and thus is a good fit for this type of data. On the Ekman-6 dataset, the supervisions have been labeled for video segments instead of individual frames. Thus, not every frame in the emotional segment necessarily expresses the emotion. This is likely a reason why the frame-based SVM performs poorly in those conditions.  

On the two-class Ekman-6 condition, BEAC-Net beats the rest, except for the very first tIoU setting. On the full Ekman-6 condition, BEAC-Net still outperforms the baselines, but the performance gap is smaller. This agrees with our observation that the full Ekman-6 is the most difficult dataset. 

We also observe that this is still a remarkable disparity among the performance of our network across the two datasets with varying complexity. Fig.~\ref{fig:attri_across} demonstrates the result. We observe that once the mAPs drop below 0.9, it would become extremely sensitive to tIoU threshold. Thus, we find it necessary to compare the performance across tIoU thresholds. 

\vspace{0.05in}

\noindent \textbf{Error Analysis.} We analyze the relation between the presence of human faces and the emotion classification accuracy of BEAC-Net. First, we detect the presence of faces in the test set of the full Ekman6 dataset, which consists of 218 videos, using a highly accurate face detection algorithm.\footnote{\url{https://github.com/ageitgey/face_recognition}. The algorithm achieves 99.28\% accuracy on the Faces in the Wild dataset \cite{FacesInTheWild2007}.} Next, we calculate the proportion of frames that contain faces for each video. 
Table~\ref{tab:failure_face} shows the classification accuracy for five buckets for different levels of face appearance. 
The accuracy of BEAC-Net generally increases when fewer frames contain faces, but the highest accuracy is achieved when 20-40\% of the frames contain faces. However, the lowest accuracy, 44.4\%, is still better than the average performance of the SVM and the ITE baseline. 

The results suggest that BEAC-Net learns to use some facial information but does not use it very effectively, leading to degraded performance when 40\% or more frames contain faces. This is consistent with the task being investigated, which is about detecting the overall emotion of videos, rather than only faces or people. We hypothesize the reason is the lack of a module dedicated to faces, such as one that recognizes facial landmarks and extracts features from them. Adding such a module would improve the data efficiency for learning facial expressions and would be a promising direction for future research.

\begin{table}[t]
\caption{\label{tab:failure_face} Classification accuracy with different proportions of frames without human face in the Ekman-6 Dataset.}
\vspace{0.3cm}
\centering
\renewcommand{\arraystretch}{1.2}
\begin{tabular}{ccc} \toprule
Frames without Faces  & Data Proportion & Classification Accuracy\\
\midrule
0-40\%  & $20.6\%$  & $44.4\%$
\\
40-60\% & $18.3\%$  & $45.0\%$
\\
60-80\% & $13.8\%$  & $56.7\%$
\\
80-95\% & $13.3\%$  & $51.7\%$
\\
$>$95\% & $31.1\%$  & $54.4\%$ \\
\bottomrule
\end{tabular} 
\end{table}

\begin{figure*}[!htb] \centering
\includegraphics[scale=0.7]{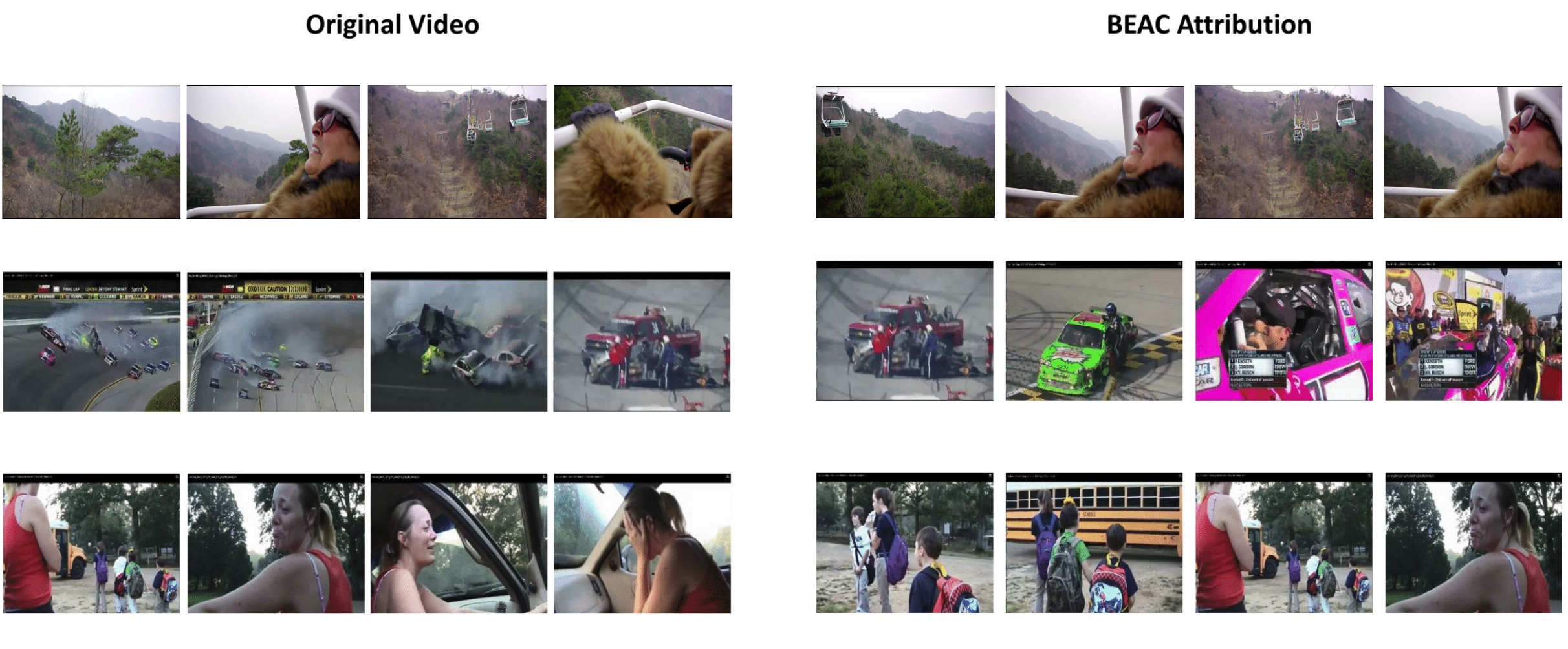} 
\caption{\label{fig:failurecase} A study of failure cases. Every row shows 4 manually extracted key frames from the original videos and from the attribution result of BEAC-Net. The correct labels for the three videos are fear, fear and sadness, respectively, but BEAC-Net classifies them as joy, sadness and anger. }
\end{figure*}

We also manually examined a few failure cases. Fig.~\ref{fig:failurecase} demonstrates the key frames in three videos where BEAC-Net failed on both the emotion attribution task and the classification task. The first case is labeled as \emph{joy} but the ground truth is \emph{fear}. The video shows a lady scared of the height on a gondola lift. As her face was mostly covered by the sunglasses and the hat, and audio information like her screaming was not used in this work, BEAC-Net was unable to recognize the emotion. 
In the second case, the emotion is expressed by a serial of  car crashes, which was perceived as fear by the annotators, but BEAC-Net's prediction of sadness is also reasonable. In the third case, BEAC-Net was not able to reconcile the emotions in the facial expression and the gesture. Holding one's head could also indicate anger, but the woman's facial expressions help resolve the ambiguity; however, BEAC-Net could not understand the two information sources jointly. 

\vspace{0.05in}
\noindent \textbf{Emotion-oriented Summarization.} We carried out a user study to quantitatively evaluate the video summaries. We randomly selected two videos from each of the 6 emotion categories in the Ekman-6 dataset and randomly assigned them to the 3-frame and the 6-frame conditions. Videos in the 3-frame condition were summarized into 3 frames and similarly for the 6-frame condition. After that, for every video, we create four summaries using three baselines and the proposed technique, yielding 48 summaries in total. The three baseline techniques are:

\noindent \textbf{Uniform}: uniformly sample the frames/clips from the videos;

\noindent \textbf{SVM}: the video is summarized by the scores of SVM prediction.
The frames with top-6 or top-3 scores of each label are selected in practice.

\noindent \textbf{ITE}: we use the summarization method based on ITE, as described in \cite{heterog_tac}.

\noindent Ten human participants rated all summaries, with no knowledge of the techniques that created them, after viewing the corresponding video.
The summaries were rated on a five-point
Likert scale and using the following four criteria \cite{heterog_tac}:

\noindent \textbf{Accuracy}: does the summary accurately describe the main content of the original video?

\noindent \textbf{Coverage}: how much content of the video is covered in the summary?

\noindent \textbf{Quality}: how is the overall subjective quality of the summary?

\noindent \textbf{Emotion}: To what extent does the summary faithfully capture the emotion in the original video? 

\begin{figure}[t] \begin{centering}
\includegraphics[scale=0.6]{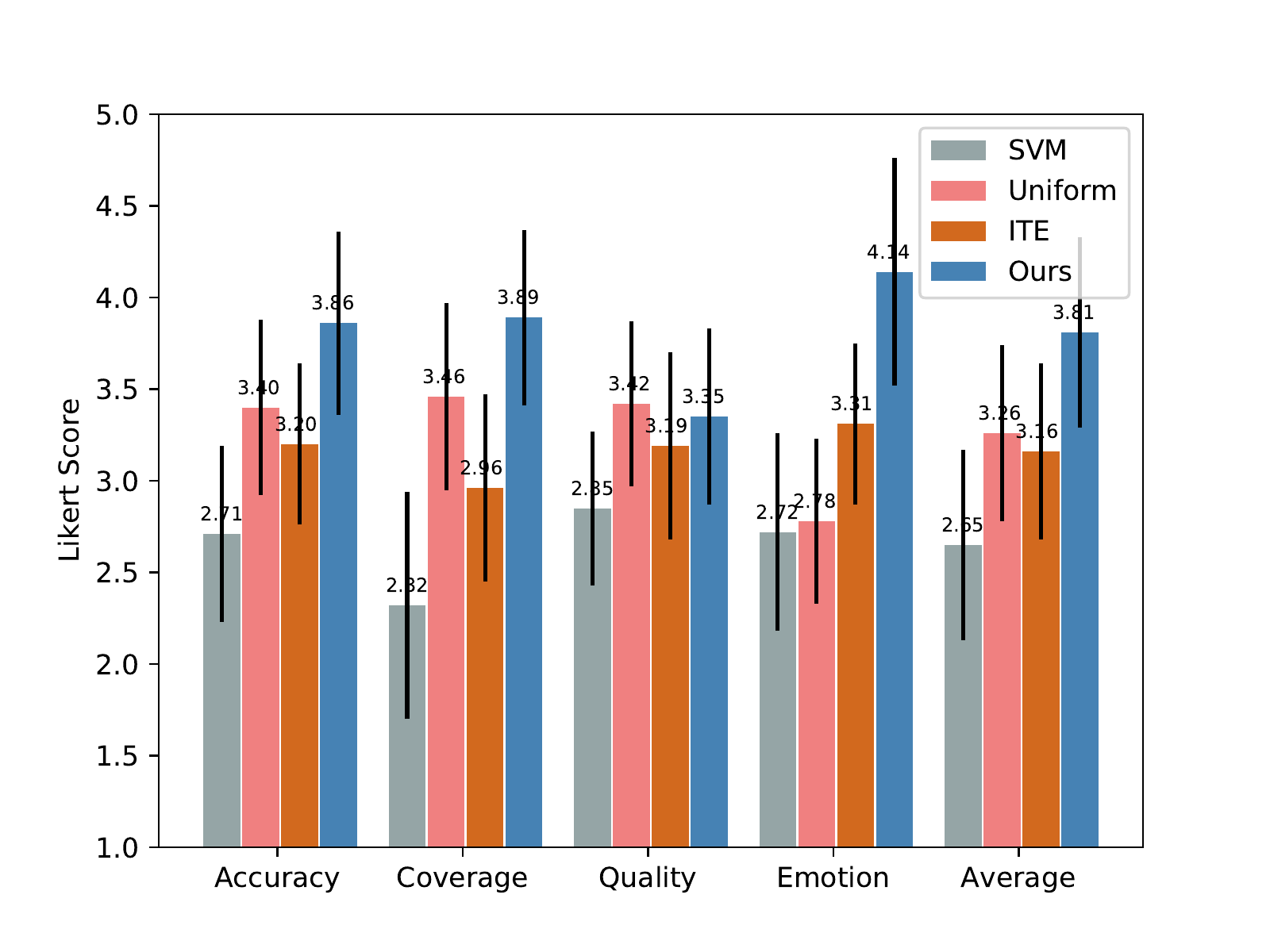} \par\end{centering}
\caption{\label{fig:Results-of-user-study} Quantitative evaluation of
video summaries, as rated by 10 human judges across 12 videos on a 1-to-5 Likert scale (higher is better). The Average column reports the average of the other four scores. The vertical black line segments indicate the standard deviation of the scores. } \end{figure}

Fig.~\ref{fig:Results-of-user-study} shows the results from the user study, where the average column reports the average rating across four questions. On four out of the five measures (including the Average), our method outperforms all baseline methods. The largest improvement of 0.83 appears on the Emotion criterion, suggesting our summaries covers emotional content substantially better than other methods. On the Quality criterion, we perform slightly worse than the uniform method, but the gap is a mere 0.07. 

\begin{figure*}[!htb]
\centering
\subfigure[]{
    \includegraphics[width=6.0in]{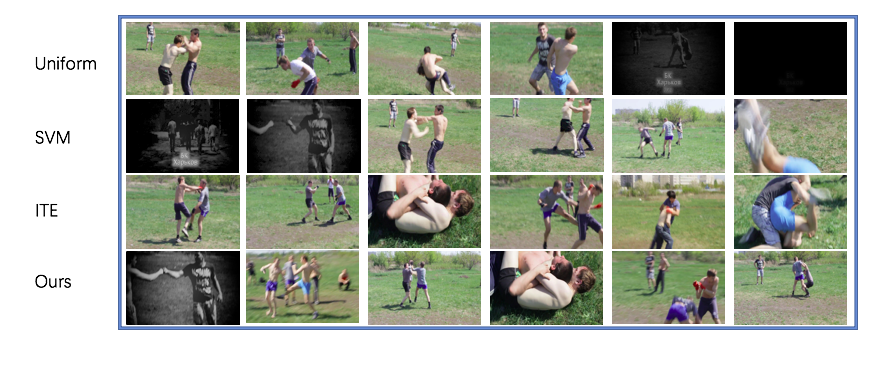}}
  \hspace{0.2in}
  \subfigure[]{
    \includegraphics[width=6.0in]{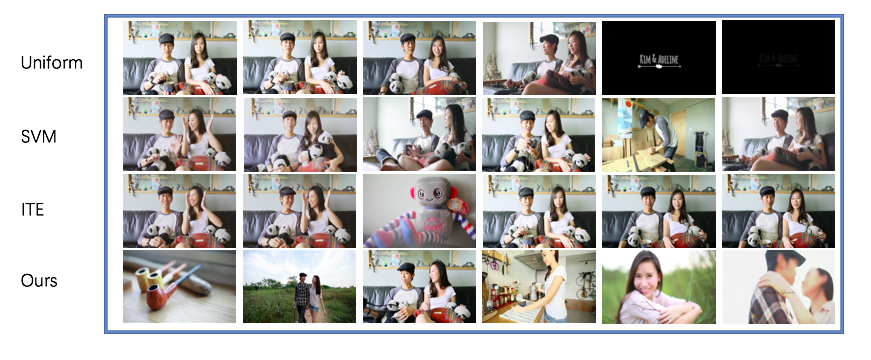}}
   \hspace{0.2in}
\subfigure[]{
    \includegraphics[width=3.0in]{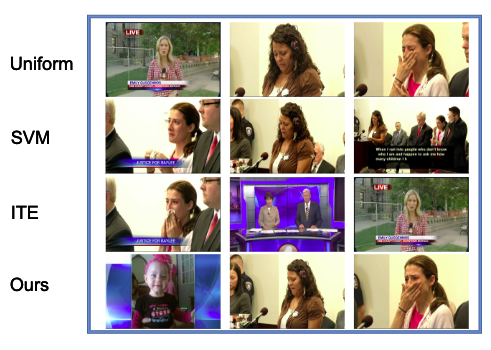}}
   \hspace{0.2in}
  \subfigure[]{
    \includegraphics[width=3.0in]{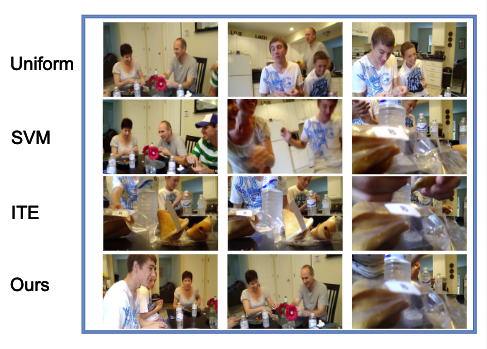}}
\caption{\label{fig:Results-of-video} Example of summaries for four videos. The top two videos are from the 6-image summary group and the bottom two are from the 3-image group. (a) shows teenagers fighting on a grass field. (b) is an interview of a young romantic couple recalling their love stories. (c) is a news report of the trial of a child murder case in the court of law.  (d) is a video showing a family party. 
} 
\end{figure*}

Figure \ref{fig:Results-of-video} compares examples from the different summarization methods. We make the following observations. First, our technique is able to cover sparsely positioned emotional expressions in the summary.  Figure \ref{fig:Results-of-video} b) contains an illustrative example. The original video contains an interview of a young romantic couple recalling their love stories. The vast majority of the video contains an interview with the couple sitting on a couch, as shown in the uniform row; emotional expressions are sparse and widely dispersed. Our model accurately chose multiple memory flashbacks as the summary while other methods give priority to the interview shots. Second, the proposed technique captures the main emotion
segment. Benefiting from the results of the attribution framework, our
summarization method focuses on clips that contain the main emotion of the video. Figure \ref{fig:Results-of-video} a) shows a video with mainly angry content, and the video summary created by our method shows the fighting
scenes.  Figure \ref{fig:Results-of-video} c) shows a video with sadness; our summary not only captures the crying but also the cause of sadness, a photo of a murdered child.

\section{Conclusions}
Computational understanding of emotions in user-generated video content is a challenging task due to the sparsity of emotional content, the presence of multiple emotions, and the variable quality of user-generated videos. We suggest that the ability to locate emotional content is crucial for accurate emotion understanding. 

Toward this end, we present a multi-task neural network with a novel bi-stream architecture, called Bi-stream Emotion Attribution-Classification Network (BEAC-Net). The network is end-to-end trainable and solves emotion recognition and attribution simultaneously. The attribution network locates the emotional content, which is processed in parallel with the original video within the bi-stream architecture. 
An ablation study shows the bi-stream architecture provides significant benefits for emotion recognition and the proposed emotion attribution network outperforms traditional temporal attention. The results indicate that the proposed technique improves the handling of sparse emotional content. 
In addition, we propose a video summarization technique based on the attribution provided by BEAC-Net. The technique outperforms existing baselines in a user study. 

Emotions play an important role in human cognition and underlie proper social interactions and daily activities. An accurate understanding of human emotions can enable many interesting applications such as story generation based on visual information \cite{PlotShot2016}. We believe this work represents a significant step in improving understanding emotional content in videos.  

\section{Acknowledgment}
This work was supported in part by 
NSFC Projects
(61572134, 61572138, U1611461),
Shanghai Sailing Program (17YF1427500), Fudan University-CIOMP Joint Fund (FC2017-006),  
STCSM Project (16JC1420400),
Shanghai Municipal Science and Technology Major Project
(2017SHZDZX01, 2018SHZDZX01) and ZJLab.

{\small{}{} \bibliographystyle{ieeetr} \bibliography{emotion_new}

\begin{thebibliography}{10}

\bibitem{Damasio1994}
A.~R. Damasio, {\em Descartes’ error: Emotion, reason and the human brain}.
\newblock New York: Avon Books, 1994.

\bibitem{Clore2009}
G.~L. Clore and J.~E. Palmer, ``Affective guidance of intelligent agents: How
  emotion controls cognition,'' {\em Cognitive Systems Research}, no.~1,
  pp.~21--30, 2009.

\bibitem{Guadagno2013}
R.~E. Guadagno, D.~M. Rempala, S.~Murphy, and B.~M. Okdie, ``What makes a video
  go viral? an analysis of emotional contagion and internet memes,'' {\em
  Computers in Human Behavior}, vol.~29, no.~6, pp.~2312--2319, 2013.

\bibitem{CAVVA}
K.~Yadati, H.~Katti, and M.~Kankanhalli, ``{CAVVA}: Computational affective
  video-in-video advertising,'' {\em IEEE Transactions on Multimedia}, vol.~16,
  no.~1, 2014.

\bibitem{Ikizler2012}
N.~Ikizler-Cinbis and S.~Sclaroff, ``Web-based classifiers for human action
  recognition,'' {\em IEEE Transactions on Multimedia}, vol.~14,
  pp.~1031--1045, Aug 2012.

\bibitem{Xu2017}
W.~Xu, Z.~Miao, X.~P. Zhang, and Y.~Tian, ``A hierarchical spatio-temporal
  model for human activity recognition,'' {\em IEEE Transactions on
  Multimedia}, vol.~19, pp.~1494--1509, July 2017.

\bibitem{Somandepalli2017}
K.~Somandepalli, N.~Kumar, T.~Guha, and S.~S. Narayanan, ``Unsupervised
  discovery of character dictionaries in animation movies,'' {\em IEEE
  Transactions on Multimedia}, vol.~PP, no.~99, pp.~1--1, 2017.

\bibitem{Joho2009}
H.~Joho, J.~M. Jose, R.~Valenti, and N.~Sebe, ``Exploiting facial expressions
  for affective video summarisation,'' in {\em Proc. ACM conference on Image
  and Video Retrieval}, 2009.

\bibitem{Zhao2011}
S.~Zhao, H.~Yao, X.~Sun, P.~Xu, X.~Liu, and R.~Ji, ``Video indexing and
  recommendation based on affective analysis of viewers,'' in {\em Proceedings
  of the 19th ACM international conference on Multimedia}, 2011.

\bibitem{Zhen2016:TMM}
Q.~Zhen, D.~Huang, Y.~Wang, and L.~Chen, ``Muscular movement model-based
  automatic {3D/4D} facial expression recognition,'' {\em IEEE Transactions on
  Multimedia}, vol.~18, pp.~1438--1450, July 2016.

\bibitem{Liu2014}
M.~Liu, S.~Shan, R.~Wang, and X.~Chen, ``Learning expressionlets on
  spatio-temporal manifold for dynamic facial expression recognition,'' in {\em
  Proceedings of the Conference on Computer Vision and Pattern Recognition},
  2014.

\bibitem{alameda2016}
X.~Alameda-Pineda, E.~Ricci, Y.~Yan, and N.~Sebe, ``Recognizing emotions from
  abstract paintings using non-linear matrix completion,'' in {\em Proceedings
  of the IEEE Conference on Computer Vision and Pattern Recognition},
  pp.~5240--5248, 2016.

\bibitem{Yazdani2011}
A.~Yazdani, K.~Kappeler, and T.~Ebrahimi, ``Affective content analysis of music
  video clips,'' in {\em Proc. 1st ACM workshop Music information retrieval
  with user-centered and multimodal strategies}, 2011.

\bibitem{heterog_tac}
B.~Xu, Y.~Fu, Y.-G. Jiang, B.~Li, and L.~Sigal, ``Heterogeneous knowledge
  transfer in video emotion recognition, attribution and summarization,'' {\em
  IEEE Trasactions on Affective Computing}, 2017.

\bibitem{baohan2014AAAI}
Y.~Jiang, B.~Xu, and X.~Xue, ``Predicting emotions in user-generated videos,''
  in {\em The AAAI Conference on Artificial Intelligence}, 2014.

\bibitem{baohan_icmr}
B.~Xu, Y.~Fu, Y.-G. Jiang, B.~Li, and L.~Sigal, ``Video emotion recognition
  with transferred deep feature encodings,'' in {\em Indian Council of Medical
  Research}, 2016.

\bibitem{emotion_net2017}
J.~Gao, Y.~Fu, Y.-G. Jiang, and X.~Xue, ``Frame-transformer emotion
  classification network,'' in {\em Proceedings of the 2017 ACM International
  Conference on Multimedia Retrieval}, 2017.

\bibitem{Ekman1972}
P.~Ekman, ``Universals and cultural differences in facial expressions of
  emotion,'' {\em Nebrasak Symposium on Motivation}, vol.~19, pp.~207--284,
  1972.

\bibitem{Ekman1999}
P.~Ekman, ``Basic emotions,'' in {\em Handbook of Cognition and Emotion}, 1999.

\bibitem{plutchik1980emotion}
R.~Plutchik and H.~Kellerman, {\em Emotion: Theory, research and experience.
  Vol. 1, Theories of emotion}.
\newblock Academic Press, 1980.

\bibitem{Gross2002}
J.~J. Gross, ``Emotion regulation: Affective, cognitive, and social
  consequences,'' {\em Psychophysiology}, vol.~39, no.~3, p.~281–291, 2002.

\bibitem{Barrett2006}
L.~F. Barrett, ``Are emotions natural kinds?,'' {\em Perspectives on
  Psychological Science}, vol.~1, no.~1, pp.~28--58, 2006.

\bibitem{Lindquist2013}
K.~A. Lindquist, E.~H. Siegel, K.~S. Quigley, and L.~F. Barrett, ``The
  hundred-year emotion war: Are emotions natural kinds or psychological
  constructions? comment on {Lench}, {Flores}, and {Bench} (2011),'' {\em
  Psychological Bulletin}, no.~1, p.~255–263, 2013.

\bibitem{Nummenmaa2013}
L.~Nummenmaa, E.~Glerean, R.~Hari, and J.~K. Hietanen, ``Bodily maps of
  emotions,'' {\em Proceedings of the National Academy of Sciences of the
  United States of America}, vol.~111, no.~2, pp.~646--651, 2013.

\bibitem{Li:Humor2015}
B.~Li, ``A dynamic and dual-process theory of humor,'' in {\em The 3rd Annual
  Conference on Advances in Cognitive Systems}, pp.~57--74, 2015.

\bibitem{TaoChen2014Deepsentibank}
T.~Chen, D.~Borth, Darrell, and S.-F. Chang, ``Deep{S}enti{B}ank: Visual
  sentiment concept classification with deep convolutional neural networks,''
  {\em CoRR}, 2014.

\bibitem{dimenaffects1980}
A.~Russell, James, ``A circumplex model of affect,'' {\em Journal of
  Personality and Social Psychology}, vol.~39, no.~6, pp.~1161--1178, 1980.

\bibitem{fontaine2007world}
J.~R. Fontaine, K.~R. Scherer, E.~B. Roesch, and P.~C. Ellsworth, ``The world
  of emotions is not two-dimensional,'' {\em Psychological Science}, vol.~18,
  no.~12.

\bibitem{lovheim2012new}
H.~L{\"o}vheim, ``A new three-dimensional model for emotions and monoamine
  neurotransmitters,'' {\em Medical Hypotheses}, vol.~78, no.~2, pp.~341--348,
  2012.

\bibitem{AVEC2017challenge1}
S.~Chen, Q.~Jin, J.~Zhao, and S.~Wang, ``Multimodal multi-task learning for
  dimensional and continuous emotion recognition,'' in {\em Proceedings of the
  7th Annual Workshop on Audio/Visual Emotion Challenge}, pp.~19--26, 2017.

\bibitem{AVEC2017challenge2}
J.~Huang, Y.~Li, J.~Tao, Z.~Lian, Z.~Wen, M.~Yang, and J.~Yi, ``Continuous
  multimodal emotion prediction based on long short term memory recurrent
  neural network,'' in {\em AVEC'17 Proceedings of the 7th Annual Workshop on
  Audio/Visual Emotion Challenge}, pp.~11--18, 2017.

\bibitem{baveye2015liris}
Y.~Baveye, E.~Dellandrea, C.~Chamaret, and L.~Chen, ``Liris-accede: A video
  database for affective content analysis,'' {\em IEEE Transactions on
  Affective Computing}, vol.~6, no.~1, pp.~43--55, 2015.

\bibitem{Benini2011}
S.~Benini, L.~Canini, and R.~Leonardi, ``A connotative space for supporting
  movie affective recommendation,'' {\em IEEE Transactions on Multimedia},
  vol.~13, no.~6, pp.~1356--1370, 2011.

\bibitem{Machajdik2010}
J.~Machajdik and A.~Hanbury, ``Affective image classication using features
  inspired by psychology and art theory,'' in {\em Proceedings of the 18th ACM
  international conference on Multimedia}, pp.~83--92, 2010.

\bibitem{Lu2012}
X.~Lu, P.~Suryanarayan, R.~B. Adams, J.~Li, M.~G. Newman, and J.~Z. Wang, ``On
  shape and the computability of emotions,'' in {\em Proceedings of the 20th
  ACM international conference on Multimedia}, 2012.

\bibitem{Jou2014}
B.~Jou, S.~Bhattacharya, and S.-F. Chang, ``Predicting viewer perceived
  emotions in animated {GIF}s,'' in {\em Proceedings of the 22nd ACM
  international conference on Multimedia}, 2014.

\bibitem{MI-SC}
W.~Hu, X.~Ding, B.~Li, J.~Wang, Y.~Gao, F.~Wang, and S.~Maybank,
  ``Multi-perspective cost-sensitive context-aware multi-instance sparse coding
  and its application to sensitive video recognition,'' {\em IEEE Transactions
  on Multimedia}, vol.~18, no.~1, 2016.

\bibitem{Song2013}
Y.~Song, L.-P. Morency, and R.~Davis, ``Learning a sparse codebook of facial
  and body microexpressions for emotion recognition,'' in {\em Proceedings of
  the 15th ACM International conference on multimodal interaction}, 2013.

\bibitem{schuller2003hidden}
B.~Schuller, G.~Rigoll, and M.~Lang, ``Hidden markov model-based speech emotion
  recognition,'' in {\em Proceedings of the 2003 International Conference on
  Multimedia and Expo - Volume 2}, ICME '03, pp.~401--404, IEEE Computer
  Society, 2003.

\bibitem{mao2014learning}
Q.~Mao, M.~Dong, Z.~Huang, and Y.~Zhan, ``Learning salient features for speech
  emotion recognition using convolutional neural networks,'' {\em IEEE
  Transactions on Multimedia}, vol.~16, no.~8, pp.~2203--2213, 2014.

\bibitem{Zhang2018}
S.~Zhang, S.~Zhang, T.~Huang, and W.~Gao, ``Speech emotion recognition using
  deep convolutional neural network and discriminant temporal pyramid
  matching,'' {\em IEEE Transactions on Multimedia}, vol.~20, no.~6,
  pp.~1576--1590, 2018.

\bibitem{Wang2006}
H.-L. Wang and L.-F. Cheong, ``Affective understanding in film,'' {\em IEEE
  TCSVT}, 2006.

\bibitem{zeng2007audio}
Z.~Zeng, J.~Tu, M.~Liu, T.~S. Huang, B.~Pianfetti, D.~Roth, and S.~Levinson,
  ``Audio-visual affect recognition,'' {\em IEEE Transactions on multimedia},
  vol.~9, no.~2, pp.~424--428, 2007.

\bibitem{acar2016comprehensive}
E.~Acar, F.~Hopfgartner, and S.~Albayrak, ``A comprehensive study on mid-level
  representation and ensemble learning for emotional analysis of video
  material,'' {\em Multimedia Tools and Applications}, vol.~76, pp.~1--29,
  2016.

\bibitem{DeepMultimodalLearning}
L.~Pang, S.~Zhu, and C.-W. Ngo, ``Deep multimodal learning for affective
  analysis and retrieval,'' {\em IEEE Transactions on Multimedia}, vol.~17,
  no.~11, 2015.

\bibitem{kahou2013combining}
S.~E. Kahou, C.~Pal, X.~Bouthillier, P.~Froumenty, {\c{C}}.~G{\"u}l{\c{c}}ehre,
  R.~Memisevic, P.~Vincent, A.~Courville, Y.~Bengio, R.~C. Ferrari, {\em
  et~al.}, ``Combining modality specific deep neural networks for emotion
  recognition in video,'' in {\em Proceedings of the 15th ACM International
  conference on multimodal interaction}, pp.~543--550, ACM, 2013.

\bibitem{You2015AAAI_img_sentiment}
Q.~You, J.~Luo, H.~Jin, and J.~Yang, ``Robust image sentiment analysis using
  progressively trained and domain transferred deep networks,'' in {\em AAAI},
  2015.

\bibitem{Borth2013acmmm}
D.~Borth, R.~Ji, T.~Chen, T.~M. Breuel, and S.-F. Chang., ``Large-scale visual
  sentiment ontology and detectors using adjective noun pairs,'' in {\em
  Proceedings of the 21st ACM international conference on Multimedia}, 2013.

\bibitem{Wang-Ji2015}
S.~Wang and Q.~Ji, ``Video affective content analysis: a survey of state of the
  art methods,'' {\em IEEE Transactions on Automatic Control}, vol.~PP, no.~99,
  pp.~1--1, 2015.

\bibitem{Arifin2008}
S.~Arifin and P.~Y.~K. Cheung, ``Affective level video segmentation by
  utilizing the pleasure-arousal-dominance information,'' {\em IEEE
  Transactions on Multimedia}, vol.~10, no.~7, 2008.

\bibitem{Truong:2007:VAS:1198302.1198305}
B.~T. Truong and S.~Venkatesh, ``Video abstraction: A systematic review and
  classification,'' {\em ACM Transactions on Multimedia Computing,
  Communications, and Applications}, vol.~3, no.~1, pp.~79--82, 2007.

\bibitem{Ma:2002:UAM:641007.641116}
Y.-F. Ma, L.~Lu, H.-J. Zhang, and M.~Li, ``A user attention model for video
  summarization,'' in {\em Proceedings of the 10th ACM international conference
  on Multimedia}, 2002.

\bibitem{lai2012key}
J.-L. Lai and Y.~Yi, ``Key frame extraction based on visual attention model,''
  {\em Journal of Visual Communication and Image Representation}, vol.~23,
  no.~1, pp.~114--125, 2012.

\bibitem{event_driven_summary}
M.~Wang, R.~Hong, G.~Li, Z.-J. Zha, S.~Yan, and T.-S. Chua, ``Event driven web
  video summarization by tag localization and key-shot identification,'' {\em
  IEEE Transactions on Multimedia}, vol.~14, no.~4, pp.~975--985, 2012.

\bibitem{WangNgo2012}
F.~Wang and C.~W. Ngo, ``Summarizing rushes videos by motion, object, and event
  understanding,'' {\em IEEE Transactions on Multimedia}, vol.~14, pp.~76--87,
  Feb 2012.

\bibitem{DBLP:conf/mm/WangJCGDW14}
X.~Wang, Y.~Jiang, Z.~Chai, Z.~Gu, X.~Du, and D.~Wang, ``Real-time
  summarization of user-generated videos based on semantic recognition,'' in
  {\em Proceedings of the 22nd ACM international conference on Multimedia},
  2014.

\bibitem{jaderberg2015spatial}
M.~Jaderberg, K.~Simonyan, A.~Zisserman, and k.~kavukcuoglu, ``Spatial
  transformer networks,'' in {\em Advances in Neural Information Processing
  Systems 28}, pp.~2017--2025, 2015.

\bibitem{singh2016end}
K.~K. Singh and Y.~J. Lee, ``End-to-end localization and ranking for relative
  attributes,'' in {\em European Conference on Computer Vision}, pp.~753--769,
  Springer, 2016.

\bibitem{show2015tell}
X.~Kelvin, L.~B. Jimmy, K.~Ryan, C.~Kyunghyun, C.~Aaron, S.~Ruslan, R.~S.
  Zemel, and B.~Yoshua, ``Show, attend, tell: Neural image caption generation
  with visual attention,'' {\em International Conference on Machine Learning},
  vol.~37, pp.~2048--2057, 2015.

\bibitem{Lin2017}
C.-H. Lin and S.~Lucey, ``Inverse compositional spatial transformer networks,''
  in {\em Proceedings of the Conference on Computer Vision and Pattern
  Recognition}, 2017.

\bibitem{simonyan2014two}
K.~Simonyan and A.~Zisserman, ``Two-stream convolutional networks for action
  recognition in videos,'' in {\em Advances in neural information processing
  systems}, pp.~568--576, 2014.

\bibitem{carreira2017quo}
J.~Carreira and A.~Zisserman, ``Quo vadis, action recognition? a new model and
  the kinetics dataset,'' in {\em Proceedings of the Conference on Computer
  Vision and Pattern Recognition}, pp.~4724--4733, 2017.

\bibitem{minmax}
Z.~Li, G.~M. Schuster, and A.~K. Katsaggelos, ``Minmax optimal video
  summarization,'' {\em IEEE Transactions on Circuits and Systems for Video
  Technology}, 2005.

\bibitem{emotion6}
K.-C. Peng, T.~Chen, A.~Sadovnik, and A.~Gallagher, ``A mixed bag of emotions:
  Model, predict, and transfer emotion distributions,'' pp.~860--868, 06 2015.

\bibitem{kingma2014adam}
D.~Kingma and J.~Ba, ``Adam: A method for stochastic optimization,'' {\em arXiv
  preprint arXiv:1412.6980}, 2014.

\bibitem{AlexNet2012}
A.~Krizhevsky, I.~Sutskever, and G.~E. Hinton, ``Imagenet classification with
  deep convolutional neural networks,'' in {\em NIPS}, 2012.

\bibitem{Anderson2018}
P.~Anderson, X.~He, C.~Buehler, D.~Teney, M.~Johnson, S.~Gould, and L.~Zhang,
  ``Bottom-up and top-down attention for image captioning and visual question
  answering,'' in {\em Proceedings of the Conference on Computer Vision and
  Pattern Recognition}, 2018.

\bibitem{LiYao2015}
L.~Yao, A.~Torabi, K.~Cho, N.~Ballas, C.~Pal, H.~Larochelle, and A.~Courville,
  ``Describing videos by exploiting temporal structure,'' in {\em Proceedings
  of the IEEE International Conference on Computer Vision}, pp.~4507--4515,
  2015.

\bibitem{caba2015activitynet}
F.~Caba~Heilbron, V.~Escorcia, B.~Ghanem, and J.~Carlos~Niebles, ``Activitynet:
  A large-scale video benchmark for human activity understanding,'' in {\em
  Proceedings of the IEEE Conference on Computer Vision and Pattern
  Recognition}, pp.~961--970, 2015.

\bibitem{FacesInTheWild2007}
G.~B.~Huang, M.~Mattar, T.~Berg, and E.~Learned-Miller, ``Labeled faces in the
  wild: A database forstudying face recognition in unconstrained
  environments,'' tech. rep., 2008.

\bibitem{PlotShot2016}
R.~Cardona-Rivera and B.~Li, ``Plotshot: Generating discourse-constrained
  stories around photos,'' in {\em Proceedings of the 12th AAAI Conference on
  Artificial Intelligence and Interactive Digital Entertainment}, 2016.

\end{thebibliography}
 }{\small \par}

\begin{IEEEbiography}[{\includegraphics[width=1in,
height=1.25in,clip, keepaspectratio]{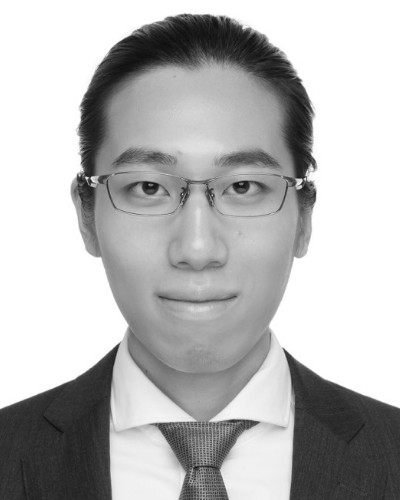}}]{Guoyun Tu} 
received his Bachelor's degree in physics at Fudan University in 2018 and
is now a graduate candidate at EIT Digital Program (Eindhoven University of Technology \& KTH Royal Institute of Technology). His research interests includes machine learning theory  and its application.
\end{IEEEbiography}

\vspace{0.2in}

 \begin{IEEEbiography}[{\includegraphics[width=1in,
height=1.25in,clip, keepaspectratio]{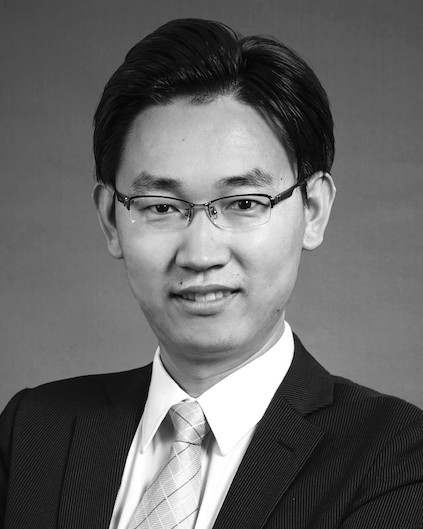}}]{Yanwei Fu} 
received the Ph.D. degree from Queen Mary University of London in 2014, and the M.Eng. degree from the Department of Computer Science and Technology, Nanjing University, China, in 2011. He held a post-doctoral position at Disney Research, Pittsburgh, PA, USA, from 2015 to 2016. He is currently a tenure-track Professor with Fudan University. His research interests are image and video understanding, and life-long learning.
\end{IEEEbiography}

\vspace{0.2in}

\begin{IEEEbiography}[{\includegraphics[width=1in,
height=1.25in,clip, keepaspectratio]{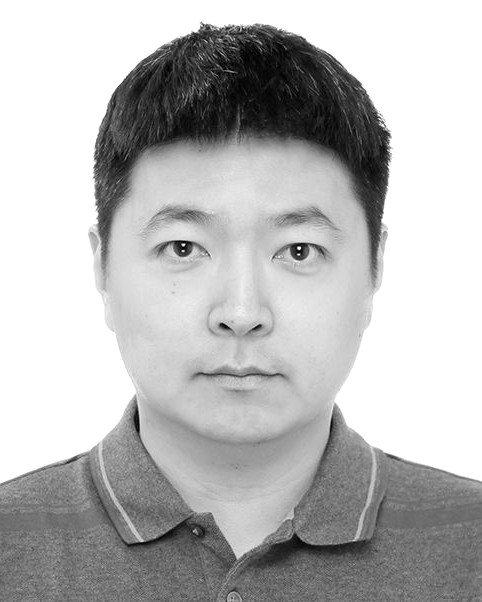}}]{Boyang Li} 
is a Senior Research Scientist at Baidu Research at Sunnyvale, Carlifornia. Prior to Baidu, he directed the Narrative Intelligence research group at Disney Research Pittsburgh. His research interests lie broadly in machine learning and multimodal reasoning, and particularly in computational understanding and generation of content with complex semantic structures, such as narratives, human emotions, and the interaction between visual and textual information. He received his Ph.D. in Computer Science from Georgia Institute of Technology in 2014, and his B. Eng. from Nanyang Technological University, Singapore in 2008. He has authored and co-authored more than 40 peer-reviewed papers in international journals and conferences.
\end{IEEEbiography}

\vspace{0.2in}

\begin{IEEEbiography}[{\includegraphics[width=1in,
height=1.25in,clip, keepaspectratio]{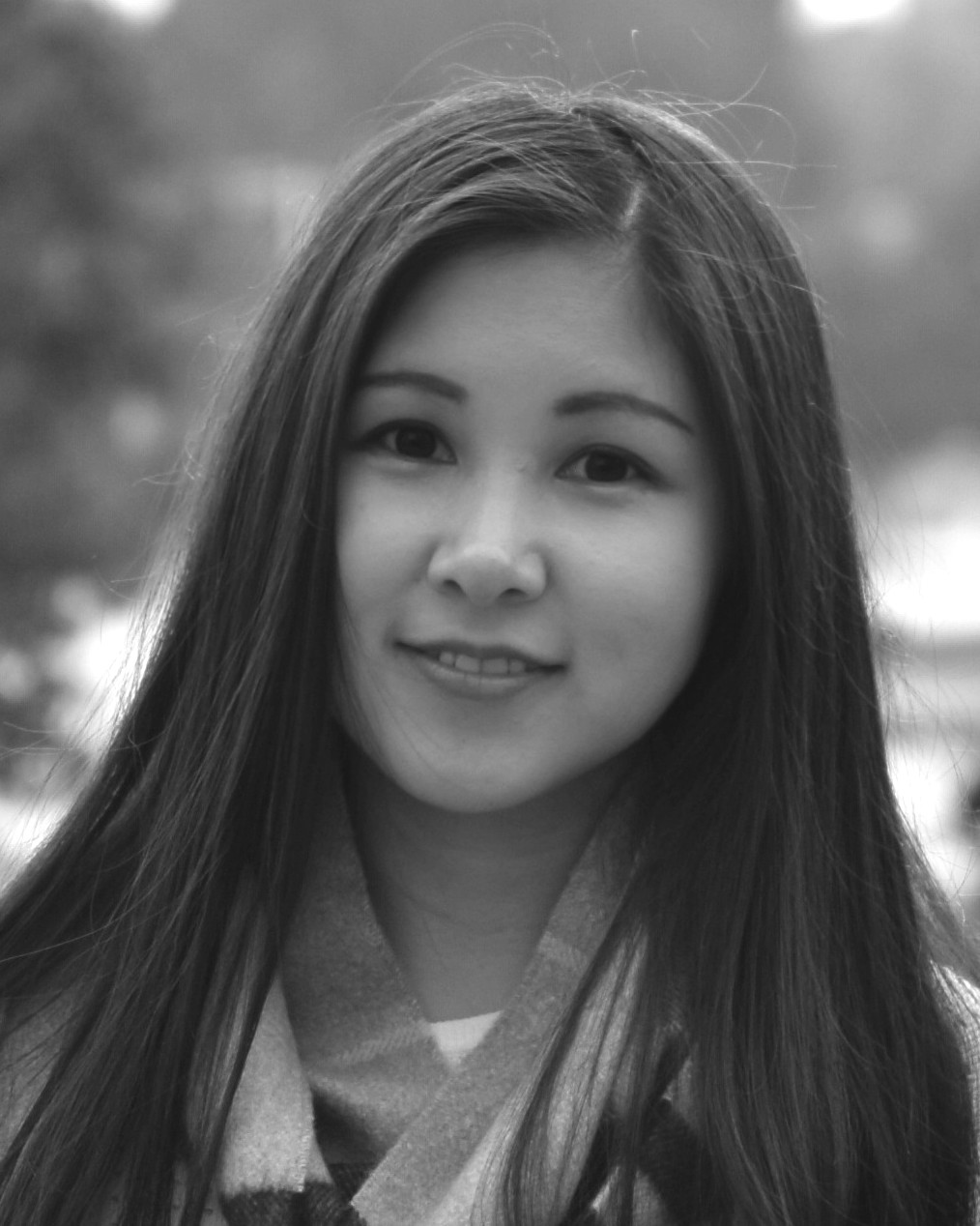}}]{Jiarui Gao}
is a first year master student majoring in software engineering at Carnegie Mellon University, Silicon Valley. She received her bachelor's degree in computer science from Fudan University, Shanghai. She was advised by Prof. Yanwei Fu and Prof. Yu-Gang Jiang to work on machine learning and multimedia then. She is passionate about full-stack software development.
\end{IEEEbiography}

\vspace{0.2in}

\begin{IEEEbiography}[{\includegraphics[width=1in,
height=1.25in,clip,keepaspectratio]{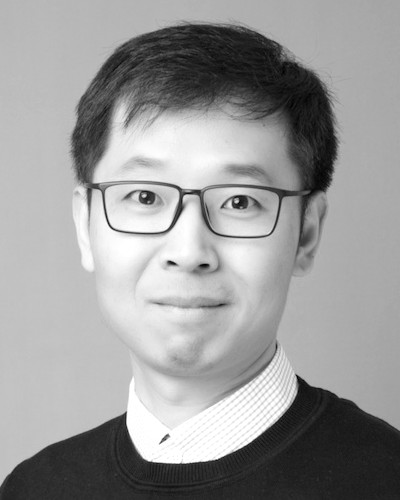}}]{Yu-Gang Jiang} is a Professor of Computer Science at Fudan University and Director of Fudan-Jilian Joint Research Center on Intelligent Video Technology, Shanghai, China. He is interested in all aspects of extracting high-level information from big video data, such as video event recognition, object/scene recognition and large-scale visual search. His work has led to many awards, including the inaugural ACM China Rising Star Award, the 2015 ACM SIGMM Rising Star Award, and the research award for outstanding young researchers from NSF China. He is currently an associate editor of ACM TOMM, Machine Vision and Applications (MVA) and Neurocomputing. He holds a PhD in Computer Science from City University of Hong Kong and spent three years working at Columbia University before joining Fudan in 2011.
\end{IEEEbiography}

\vspace{0.2in}

\begin{IEEEbiography}[{\includegraphics[width=1in,
height=1.25in,clip,keepaspectratio]{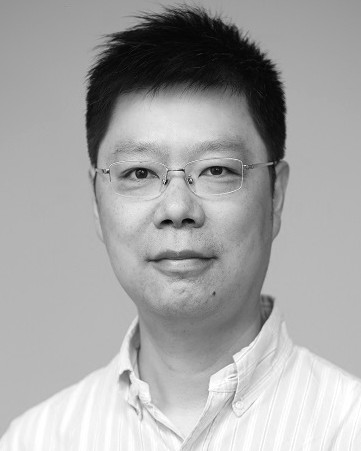}}]{Xiangyang Xue}received the BS, MS, and PhD degrees in communication engineering from Xidian University, Xi'an, China, in 1989, 1992, and 1995, respectively. He is currently a professor of computer science with Fudan University, Shanghai, China. His research interests include computer vision, multimedia information processing and machine learning.
\end{IEEEbiography}

\end{document}